%% file: arxiv.tex
\documentclass[10pt,twocolumn,letterpaper]{article}

\usepackage{cvpr}
\usepackage{times}
\usepackage{epsfig}
\usepackage{graphicx}
\usepackage{amsmath}
\usepackage{amssymb}
\usepackage{multirow}
\usepackage{tabularx}
\usepackage{ marvosym }
\usepackage{mathtools}
\usepackage{bbm}
\usepackage[export]{adjustbox}
\usepackage{booktabs}       %
\usepackage{subcaption}
\usepackage{wrapfig}
\graphicspath{{images/}{images}}
\input{defines}
\usepackage[dvipsnames]{xcolor}
\usepackage[title]{appendix}
\usepackage[pagebackref=true,breaklinks=true,letterpaper=true,colorlinks,bookmarks=false]{hyperref}

 \cvprfinalcopy %
\newcommand{\authSpace}{\quad\quad}

\def\cvprPaperID{3689} %

\newcommand{\myparagraph}[1]{\vspace{1pt}\noindent{\textbf{#1.}}}

\newcolumntype{x}[1]{>{\centering\arraybackslash\hspace{0pt}}p{#1}}
\begin{document}

\title{Not Using the Car to See the Sidewalk --- Quantifying \\and Controlling the Effects of Context in Classification and Segmentation\vspace*{-2mm}}

\author{
  Rakshith Shetty\textsuperscript{1} \authSpace Bernt Schiele\textsuperscript{1}\authSpace Mario Fritz\textsuperscript{2}
  \\[2mm]
  \textsuperscript{1}Max Planck Institute for Informatics, Saarland Informatics Campus \\
  \textsuperscript{2}CISPA Helmholtz Center i.G.,  Saarland Informatics Campus\\
  \textsuperscript{1}\texttt{\small firstname.lastname@mpi-inf.mpg.de} \\
  \textsuperscript{2}\texttt{\small firstname.lastname@cispa.saarland}\\
}
\maketitle
\begin{abstract}
Importance of visual context in scene understanding tasks is well recognized 
in the computer vision community. However, to what extent the computer vision models for image classification and semantic segmentation are dependent on the context to make their predictions is unclear. A model overly relying on context will fail when encountering objects in context distributions different from training data and hence it is important to identify these dependencies before we can deploy the models in the real-world. We propose a method to quantify the sensitivity of black-box vision models to visual context by editing images to remove selected objects and measuring the response of the target models. We apply this methodology on two tasks, image classification and semantic segmentation, and discover undesirable dependency between objects and context, for example that ``sidewalk'' segmentation relies heavily on ``cars'' being present in the image. We propose an object removal based data augmentation solution to mitigate this dependency and increase the robustness of classification and segmentation models to contextual variations. Our experiments show that the proposed data augmentation helps these models improve the performance in out-of-context scenarios, while preserving the performance on regular data.%
\end{abstract}

\vspace*{-4mm}
\section{Introduction}

\input{intro}
\section{Related work}
\input{related}
\section{Quantifying the role of context}
\input{model}

\section{Experiments and Results}
\input{results}
\section{Conclusions}
\input{conclusions}

{\small
\bibliographystyle{ieee}
\bibliography{biblioLong,paper}
}
\begin{appendices}
\input{supplementary_material}
\end{appendices}

\end{document}

%% file: defines.tex
 \makeatletter
 \DeclareRobustCommand\onedot{\futurelet\@let@token\@onedot}
 \def\@onedot{\ifx\@let@token.\else.\null\fi\xspace}

 \makeatother

\DeclareRobustCommand{\figref}[1]{Figure~\ref{#1}}

\DeclareRobustCommand{\Figref}[1]{Figure~\ref{#1}}

\DeclareRobustCommand{\secref}[1]{Section~\ref{#1}}

\DeclareRobustCommand{\Tableref}[1]{Table~\ref{#1}}
\DeclareRobustCommand{\tableref}[1]{Table~\ref{#1}}

%% file: intro.tex
\begin{figure}
    \setlength{\tabcolsep}{1pt}
    \noindent\makebox[\linewidth]{
    \scriptsize
    \begin{tabular}{ccc}
          \includegraphics[valign = c, height=0.15\linewidth, width=0.3\linewidth]{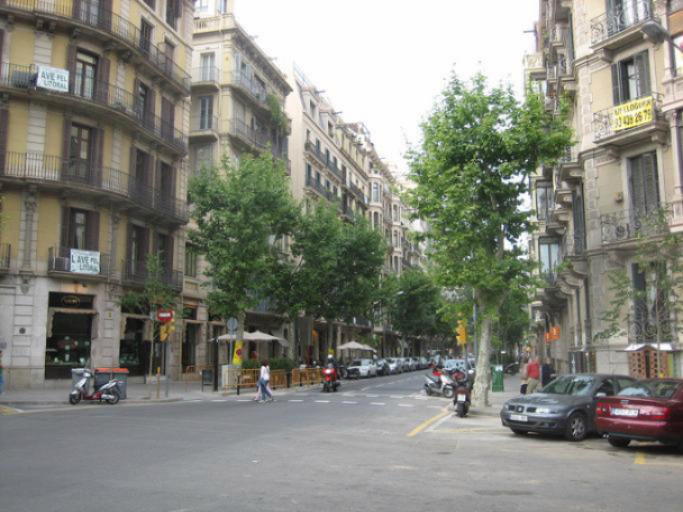}
        & \includegraphics[valign = c, height=0.15\linewidth, width=0.3\linewidth]{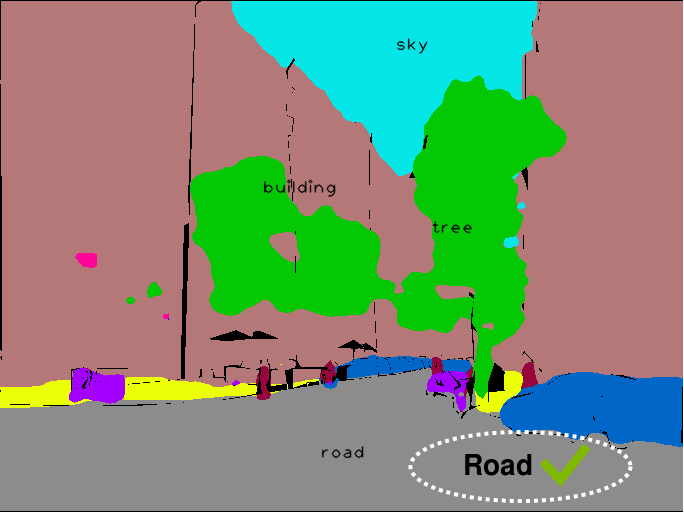} 
        & \includegraphics[valign = c, height=0.15\linewidth, width=0.3\linewidth]{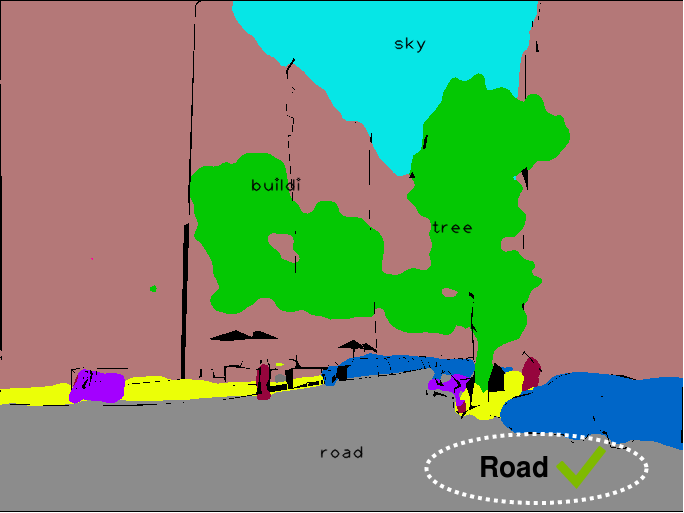} \\
        Original($\mathcal{I}$) & Upernet~\cite{xiao2018unified}  & Ours \\[0.3em]
          \includegraphics[valign = c, height=0.15\linewidth, width=0.3\linewidth]{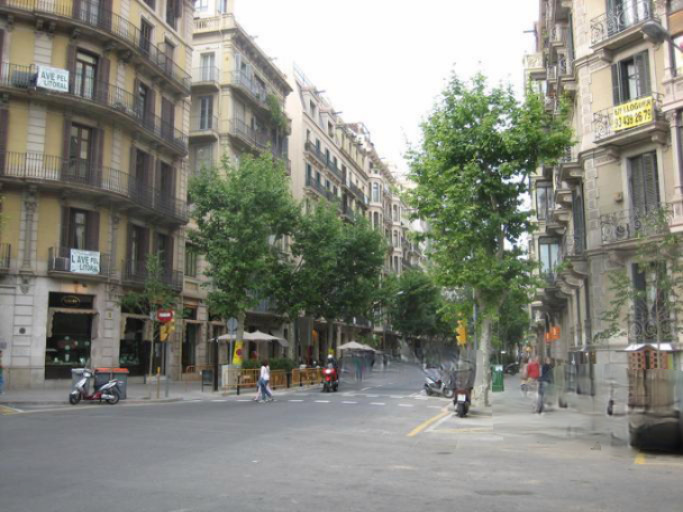}
        & \includegraphics[valign = c, height=0.15\linewidth, width=0.3\linewidth]{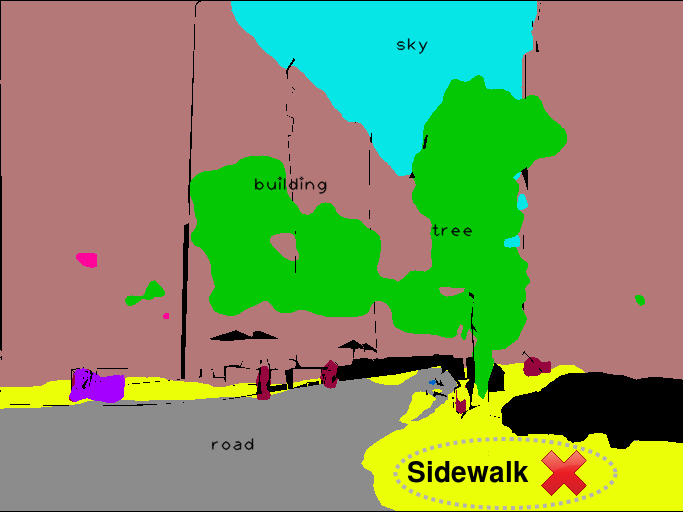} 
        & \includegraphics[valign = c, height=0.15\linewidth, width=0.3\linewidth]{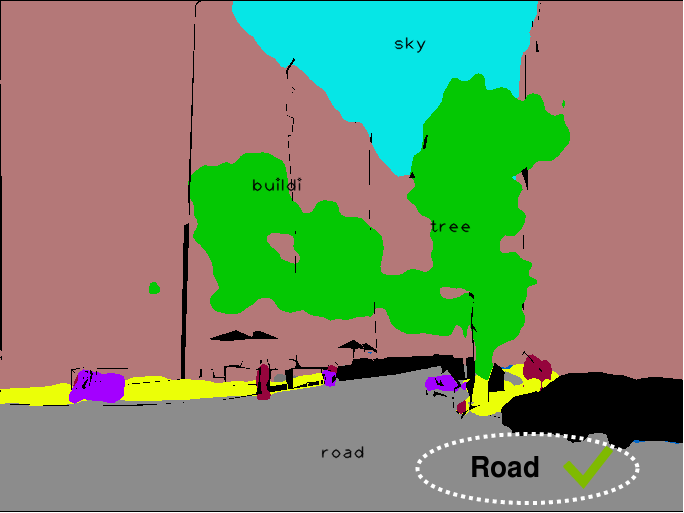} \\
        $\mathcal{I} - car$ & Upernet~\cite{xiao2018unified}  & Ours \\[0.3em] 
        \end{tabular}
        }
        \caption{An example of the sensitivity of \emph{road} and \emph{sidewalk} segmentation to the context object \emph{car}. Removing \emph{car} from the image~(second row) causes segmentation errors in the baseline model which hallucinates a sidewalk~(yellow) when there is none. Our model trained with proposed data-augmentation is more robust to these context changes.}
    \label{fig:teaser}
\end{figure}

Visual context of an object in an image is an important source of information for scene understanding tasks in both human and computer vision~\cite{torralba2010using,parikh2012exploring}. Contextual cues such as presence of frequently co-occurring objects can help resolve ambiguities between visually similar classes and improve performance in various vision tasks including object detection~\cite{mottaghi2014role,bell2016inside} and segmentation~\cite{Zhang_2018_CVPR}. 
However, objects can also appear in previously unseen context or be absent from a very typical context. For example, we might find a keyboard on a desk without a monitor~(object-without-context), or find a monitor without a keyboard~(context-without-object).
While humans can handle both these atypical scenarios gracefully, computer vision models often fail by ignoring the visual evidence for the object in object-without-context case or hallucinating objects which are not actually present in the image in context-without-object case. 
For example, in our experiments we find that \emph{keyboard} is often not recognized without a nearby \emph{monitor}, and semantic segmentation of roads suffers without \emph{cars}~(see \figref{fig:teaser}). 
While context can be an important cue, this kind of too heavy or even pathological dependency on contextual signals is undesirable, and it is important to systematically identify and ideally fix such cases.
In this work, we analyze and quantify the effect contextual information on two tasks, multi-label classification and semantic segmentation.

Context includes a lot of different kinds of information, including co-occurring objects, scene type and lighting. For our analysis, we limit context to only the set of co-occurring objects in the image. While this might seem restrictive, we find in our analysis that image classification and segmentation models learn many interesting and undesirable dependencies between an object and other co-occurring objects~(context) in the image. 
We use object removal as the main methodology to understand and quantify the role of context in downstream vision models. Specifically, we compare the output of the target models on the original input image and an edited version of this image with one object removed from it. If the model heavily uses the contextual relationship between removed object and the objects present in the image, removal will have an adverse effect on the model output. Measuring this helps us quantify the contextual dependencies learnt by the target model. %

Ideally we want models which can utilize contextual cues when available, but are robust to variations in context and can detect and segment objects even when they appear out of context. However, machine learning based vision models widely used today are biased to the data they see frequently in training and tend to perform poorly on less frequent situations, for example the object-without-context and context-without-object scenarios we are interested in.
We address this by proposing a data augmentation scheme to expose the image classification and segmentation models to different contexts during training, in-order to improve robustness of the model to context changes. This is done by removing selected objects from images and training the models on the edited images to recognize and segment other objects in the image, even with contextual objects removed. Our experiments show that the classification and segmentation models trained with this data augmentation scheme are less sensitive to context changes and perform better on real out-of-context datasets, while preserving the baseline performance on the regular data splits.  

To summarize, the main contributions  of this paper are as follows: a) We propose an object removal based method to understand and quantify sensitivity of vision models to context, b) We apply this to analyze image classification and segmentation models and find some interesting and undesirable dependencies learnt by the models between classes and contextual objects and c) We propose a data augmentation scheme based on object removal to make the models more robust to contextual variation and show that it helps improve performance in out-of-context  scenarios.

%% file: related.tex
The importance of semantic context in visual recognition is a well established with studies showing context can help humans recognize objects faster e.g. when dealing with difficult low resolution images~\cite{parikh2012exploring,barenholtz2014quantifying}. In computer vision, incorporating context information has been shown to improve performance in various tasks including object recognition~\cite{marszalek2007semantic,torralba2010using, rabinovich2007objects} and action recognition~\cite{jain201515}, object detection~\cite{bell2016inside} and segmentation~\cite{Zhang_2018_CVPR}. 
Earlier approaches built explicit context models for example by incorporating co-occurrences~\cite{rabinovich2007objects} and spatial location statistics~\cite{desai2011discriminative}. 
However, recently, explicit context modeling has been replaced the use of deep convolutional neural network~(CNN) encoders which summarize the information in the whole image into compact features. Classification and segmentation models, built on top of these deep feature encoders, can exploit information about object and context to achieve good performance~\cite{long2015fully, durand2016weldon, oquab2015object}. Approaches to improve the use of context in CNNs have been explored including using spatial pyramids~\cite{zhao2017pyramid}, atrous convolutions~\cite{chen2018deeplab} and learning context encoding with a separate neural network~\cite{Zhang_2018_CVPR}. While this implicit context encoding with deep CNNs gives good performance, it is less interpretable and it makes it hard to know whether the models are basing their decisions on visual or contextual evidence.
Recent works propose to address this with methods to interpret neural networks by visualizing salient regions for classification decision~\cite{ribeiro2016should, zeiler2014visualizing}, quantifying how interpretable individual units in network are~\cite{netdissect2017}.
While these works focus on interpreting the internal representations of the network, we look at quantifying the context sensitivity of models from the input data perspective and treat the networks as black boxes. 
By manipulating the input image to remove objects and observing the network output, we quantify the sensitivity of classification and segmentation models to context and discover some interesting and undesirable dependency between classes. A recent work \cite{rosenfeld2018elephant} takes a related approach. By adding few out-of-context objects into images they show that object detection networks are brittle to presence of out-of-context objects. However while the focus in~\cite{rosenfeld2018elephant} is to explore feature interference caused by out of context objects, we aim to quantify and mitigate contextual dependencies between classes. Recent work~\cite{dvornik2018modeling} proposes data augmentation scheme for object detection by adding objects into new contexts with contextual modeling to improve average case performance. In contrast, our work focuses on improving the contextual robustness of segmentation and classification models.%

%% file: model.tex
We quantify the contextual dependence of image classification and segmentation models by applying object removal. We propose metrics to measure the effects of context by measuring the change in the target model output for the original and the images edited to remove context objects.
Next, we will discuss our removal model, define the robustness metrics and present the data augmentation strategies to reduce the contextual dependence and improve performance in out-of-context situations.
\subsection{Object removal}
\label{sec:object_removal_model}
\begin{figure}
    \setlength{\tabcolsep}{0.1mm}
    \noindent\makebox[\linewidth]{
    \scriptsize
    \begin{tabular}{x{1.8cm}cx{5mm}c}
        & \bf\small \textcolor{blue}{Object} without \textcolor{magenta}{Context} && \bf\small \textcolor{magenta}{Context} without \textcolor{blue}{Object}\\[0.5em]
         \includegraphics[valign =c, width=0.9\linewidth]{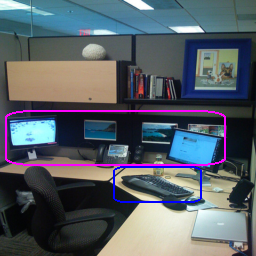} Original
        & \includegraphics[valign = c, width=0.34\linewidth]{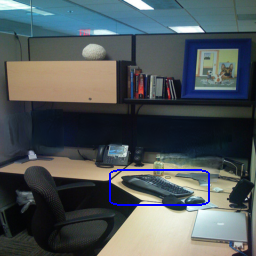}&
        & \includegraphics[valign = c, width=0.34\linewidth]{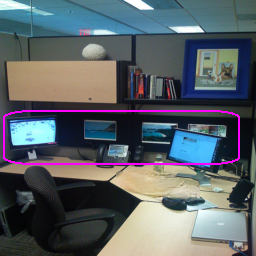}\\
         \bf Regular       & $S(\text{\it keyboard})=\textcolor{red}{1.99}$\textcolor{red}{\Lightning}         &\multirow{2}{*}{\large$\boldmath\geq$ }&$S(\text{\it keyboard)} = \textcolor{red}{4.67}$\textcolor{red}{\Lightning}\\
         \bf Ours & $S(\text{\it keyboard})=\textcolor{PineGreen}{3.40}$    &&$S(\text{\it keyboard}) = \textcolor{PineGreen}{1.39}$\\[0.5em]

         \includegraphics[valign = c, width=0.9\linewidth]{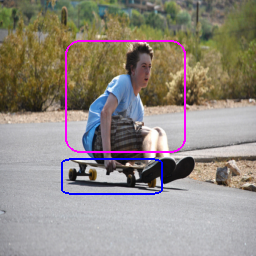} Original
        & \includegraphics[valign = c, width=0.34\linewidth]{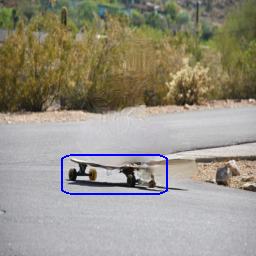}&
        & \includegraphics[valign = c, width=0.34\linewidth]{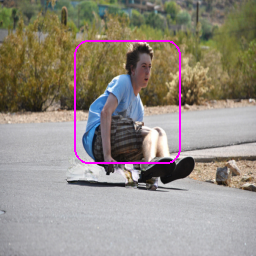}\\

         \bf Regular       & $S(\text{\it skate}) = \textcolor{red}{0.39}$\textcolor{red}{\Lightning}       &\multirow{2}{*}{\large$\boldmath\geq$ }& $S(\text{\it skate}) = \textcolor{red}{2.97}$\textcolor{red}{\Lightning}         \\
         \bf Ours & $S(\text{\it skate}) = \textcolor{PineGreen}{2.33}$ && $S(\text{\it skate}) = \textcolor{PineGreen}{-0.13}$ \\
 
         \includegraphics[valign = c, width=0.9\linewidth]{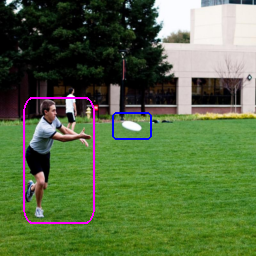} Original
        & \includegraphics[valign = c, width=0.34\linewidth]{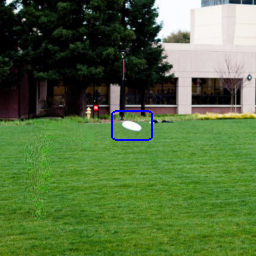}&
        & \includegraphics[valign = c, width=0.34\linewidth]{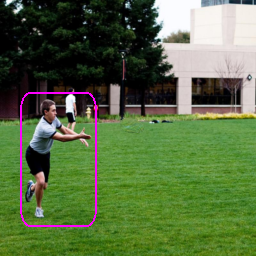}\\
         \bf Regular       & $S(\text{\it frisbee})=\textcolor{red}{0.39}$\textcolor{red}{\Lightning}    &\multirow{2}{*}{\large$\boldmath\geq$ }& $S(\text{\it frisbee}) = \textcolor{red}{2.06}$\textcolor{red}{\Lightning}\\
         \bf Ours & $S(\text{\it frisbee})=\textcolor{PineGreen}{3.32}$    && $S(\text{\it frisbee}) = \textcolor{PineGreen}{0.23}$ \\[0.5em]

         \includegraphics[valign = c, width=0.9\linewidth]{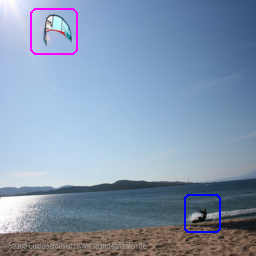} Original
        & \includegraphics[valign = c, width=0.34\linewidth]{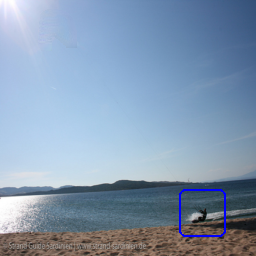}&
        & \includegraphics[valign = c, width=0.34\linewidth]{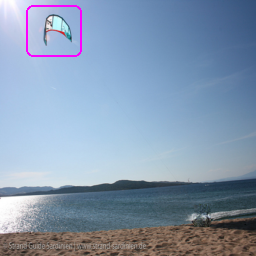}\\

         \bf Regular       & $S(\text{\it person}) = \textcolor{red}{2.15}$\textcolor{red}{\Lightning}      &\multirow{2}{*}{\large$\boldmath\geq$ }& $S(\text{\it person}) = \textcolor{red}{2.79}$\textcolor{red}{\Lightning}      \\ 
         \bf Ours & $S(\text{\it person}) = \textcolor{PineGreen}{2.83}$&& $S(\text{\it person}) = \textcolor{PineGreen}{-2.20}$\\[0.5em]

    \end{tabular}
    }
    \vspace*{-2mm}
    \caption{Context violations by image-level classifier. The primary object is marked with blue box and the context object is marked with magenta. The first column shows the original image, middle shows the image with only object and the third with only the context. We see that the baseline classifier depends heavily on the context and always scores the context only images~(last column) higher than the image with only the primary object (middle column). The data augmented model does better and gets the ordering right. }
    \vspace*{-1em}
    \label{fig:classInconsist}
\end{figure}

\begin{figure*}
    \setlength{\tabcolsep}{1pt}
    \noindent\makebox[\textwidth]{
    \scriptsize
    \begin{tabular}{cccccc}
          \includegraphics[valign = c, height=0.06\linewidth, width=0.16\linewidth]{segmentation/correlErrors/ADE_val_00000819_sidewalk/original}
        & \includegraphics[valign = c, height=0.06\linewidth, width=0.16\linewidth]{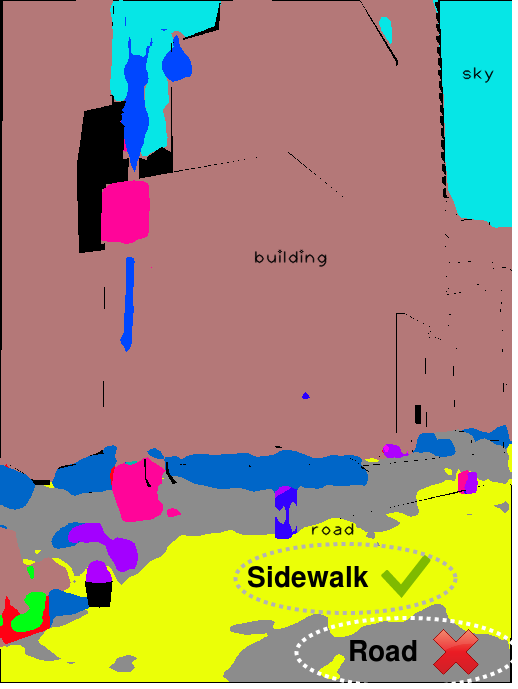} 
        & \includegraphics[valign = c, height=0.06\linewidth, width=0.16\linewidth]{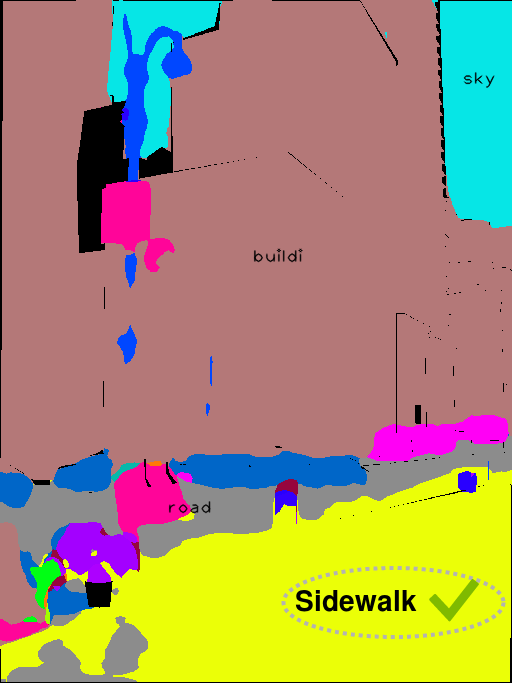} 
        & \includegraphics[valign = c, height=0.06\linewidth, width=0.16\linewidth]{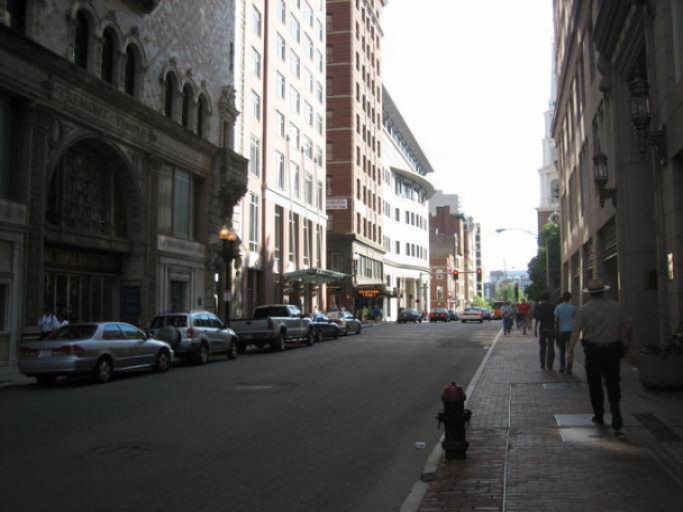}
        & \includegraphics[valign = c, height=0.06\linewidth, width=0.16\linewidth]{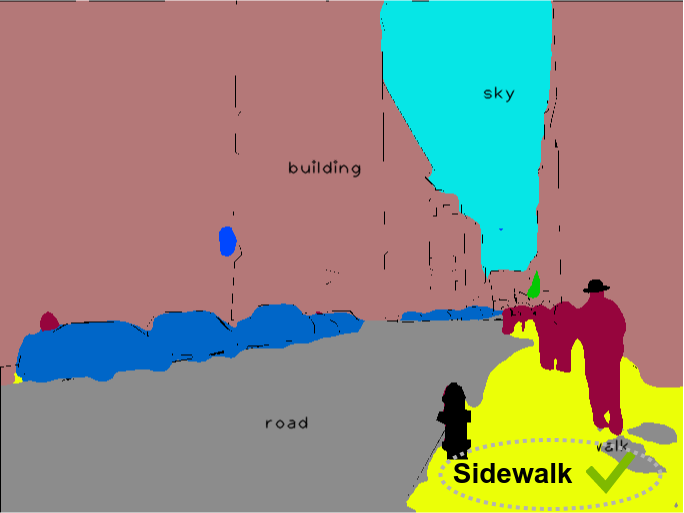}
        & \includegraphics[valign = c, height=0.06\linewidth, width=0.16\linewidth]{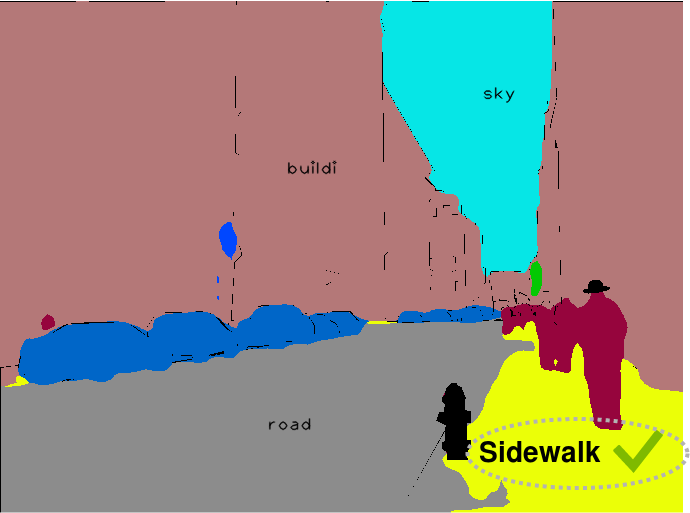}\\
          original $(\mathcal{I})$ & Upernet   & Ours & original $(\mathcal{I})$ & Upernet & Ours\\
          \includegraphics[valign = c, height=0.06\linewidth, width=0.16\linewidth]{segmentation/correlErrors/ADE_val_00000819_sidewalk/original_edited}
        & \includegraphics[valign = c, height=0.06\linewidth, width=0.16\linewidth]{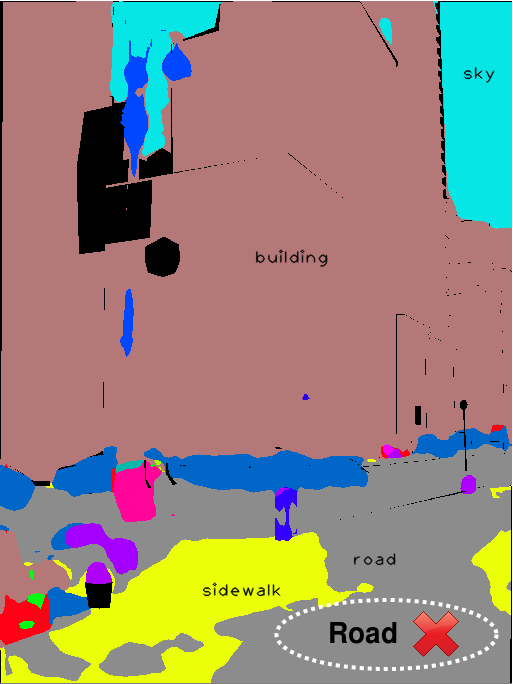}
        & \includegraphics[valign = c, height=0.06\linewidth, width=0.16\linewidth]{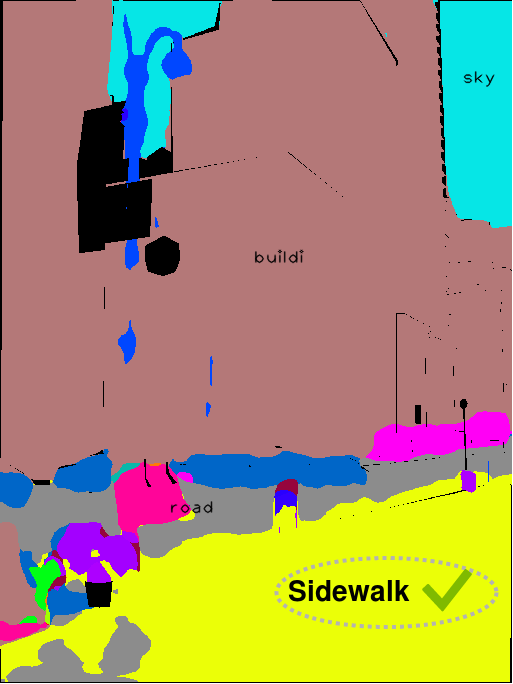}
        & \includegraphics[valign = c, height=0.06\linewidth, width=0.16\linewidth]{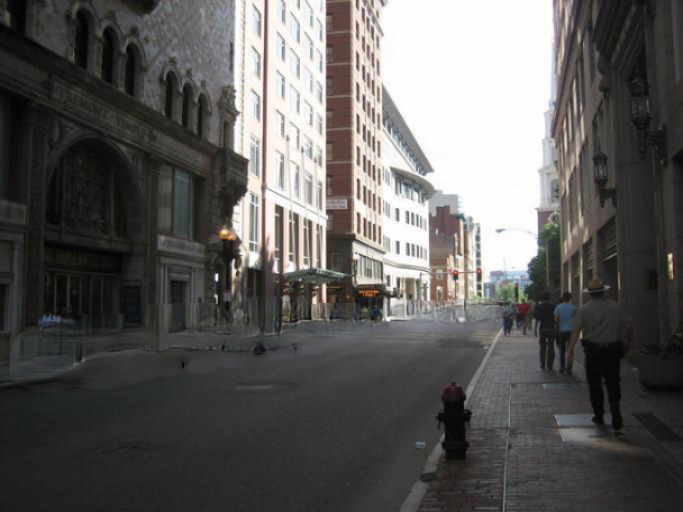}
        & \includegraphics[valign = c, height=0.06\linewidth, width=0.16\linewidth]{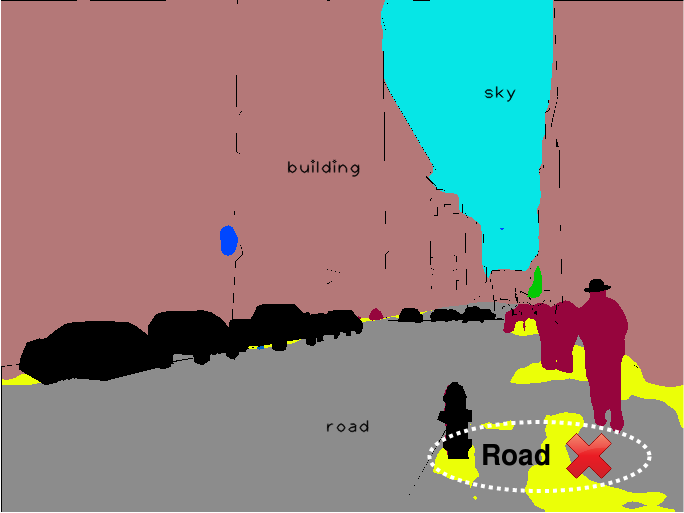}
        & \includegraphics[valign = c, height=0.06\linewidth, width=0.16\linewidth]{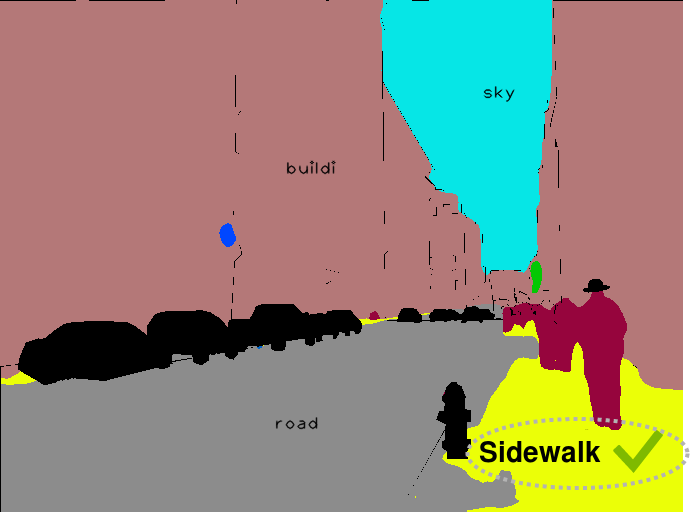}\\
          $\mathcal{I} - sign $ & Upernet & Ours & $\mathcal{I} - car$ & Upernet & Ours\\[0.3em]
        \includegraphics[valign = c, height=0.06\linewidth, width=0.16\linewidth]{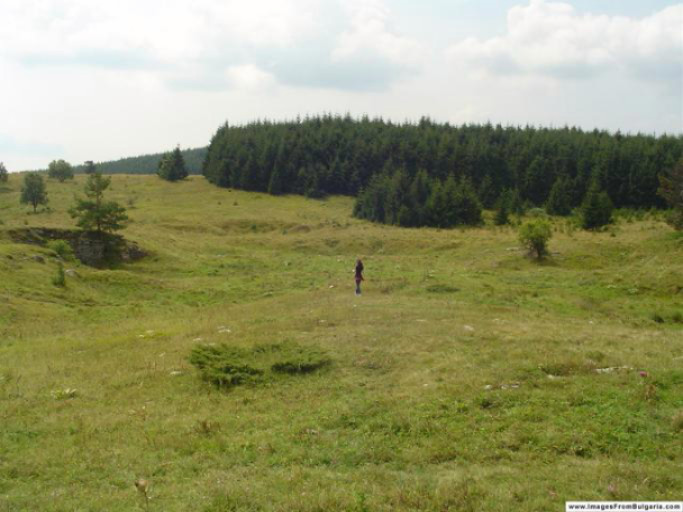}
        & \includegraphics[valign = c, height=0.06\linewidth, width=0.16\linewidth]{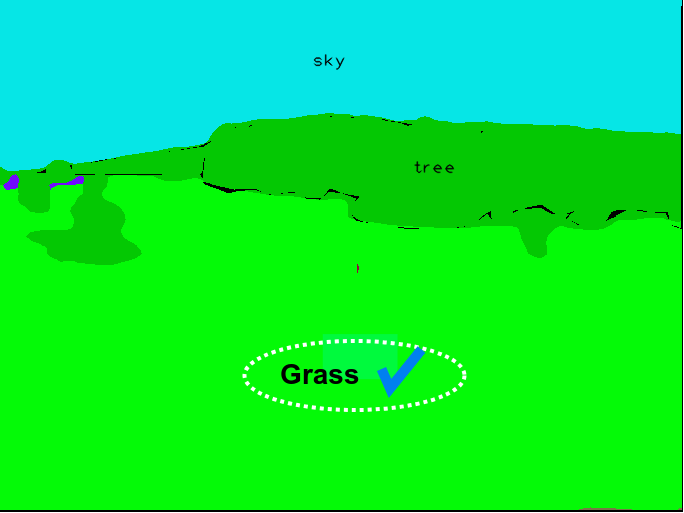} 
        & \includegraphics[valign = c, height=0.06\linewidth, width=0.16\linewidth]{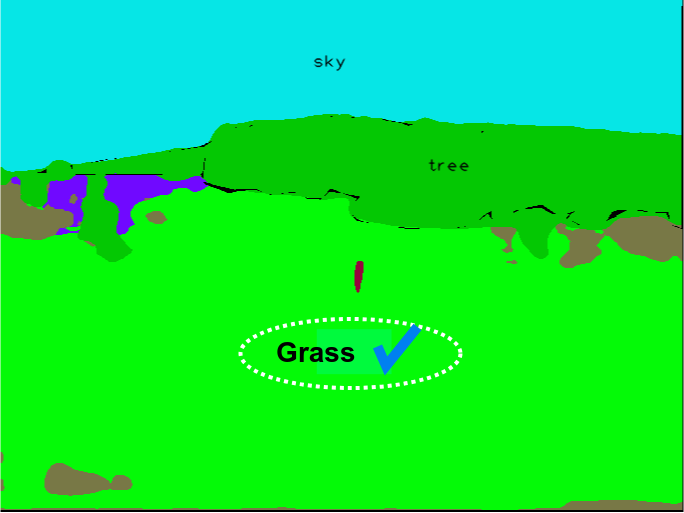} 
        & \includegraphics[valign = c, height=0.06\linewidth, width=0.16\linewidth]{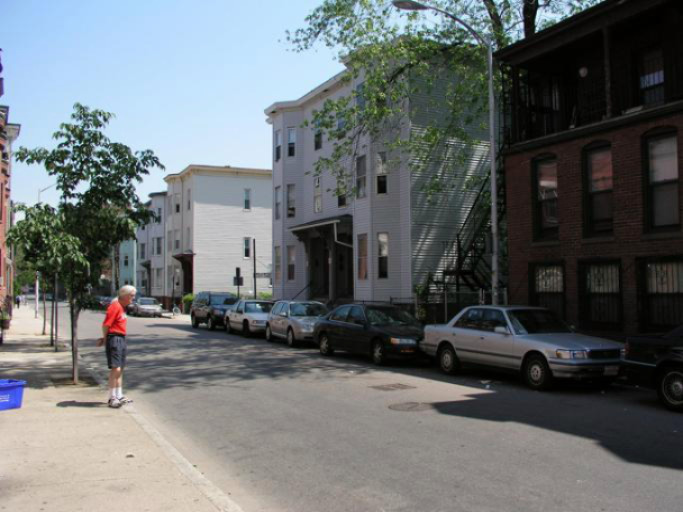}
        & \includegraphics[valign = c, height=0.06\linewidth, width=0.16\linewidth]{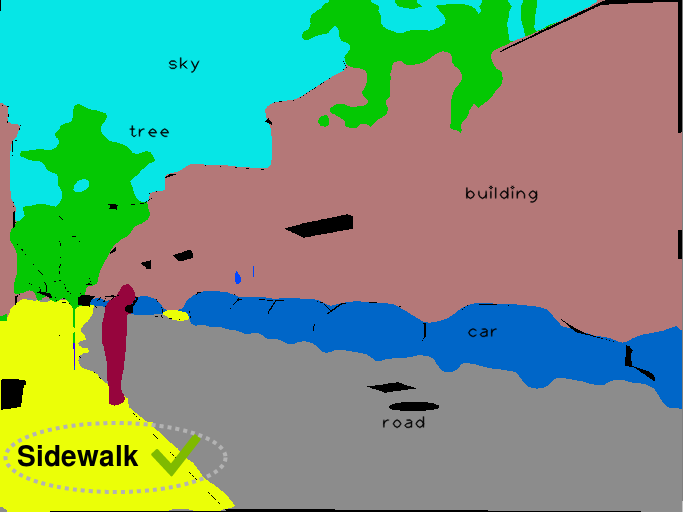}
        & \includegraphics[valign = c, height=0.06\linewidth, width=0.16\linewidth]{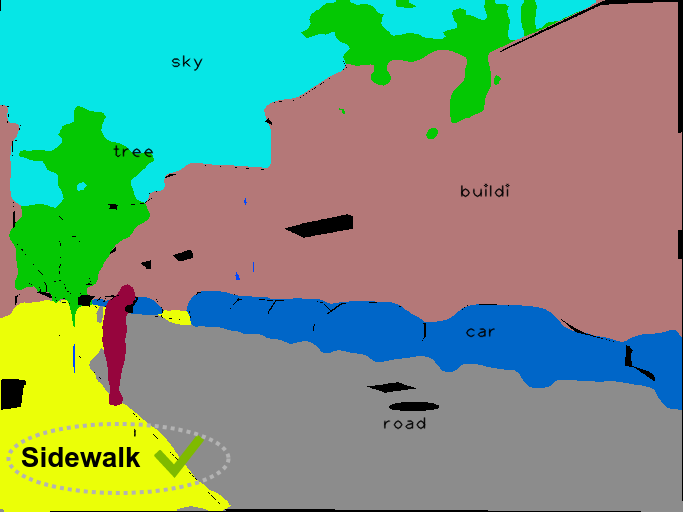}\\
          original $(\mathcal{I})$ & Upernet  & Ours & original $(\mathcal{I})$ & Upernet & Ours\\
        \includegraphics[valign = c, height=0.06\linewidth, width=0.16\linewidth]{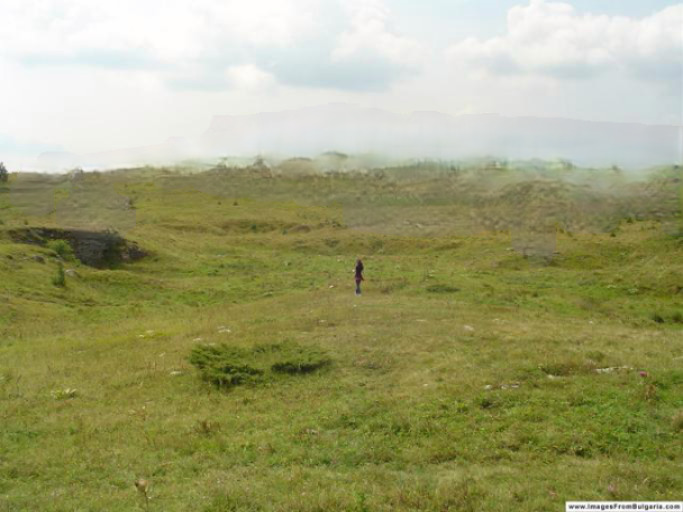}
        & \includegraphics[valign = c, height=0.06\linewidth,  width=0.16\linewidth]{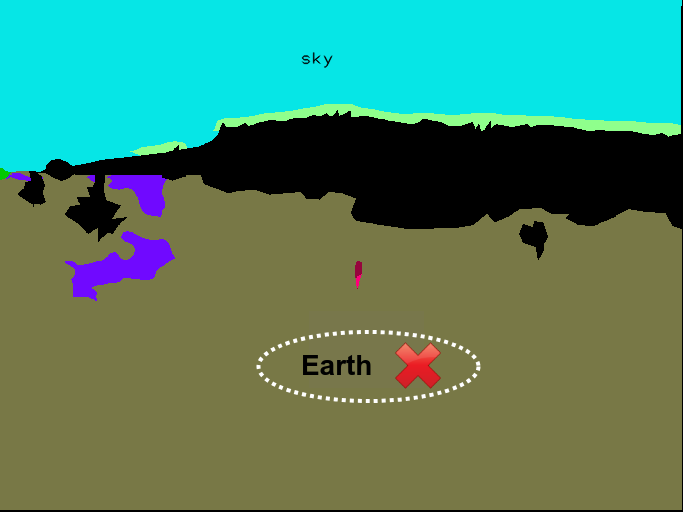} 
        & \includegraphics[valign = c, height=0.06\linewidth, width=0.16\linewidth]{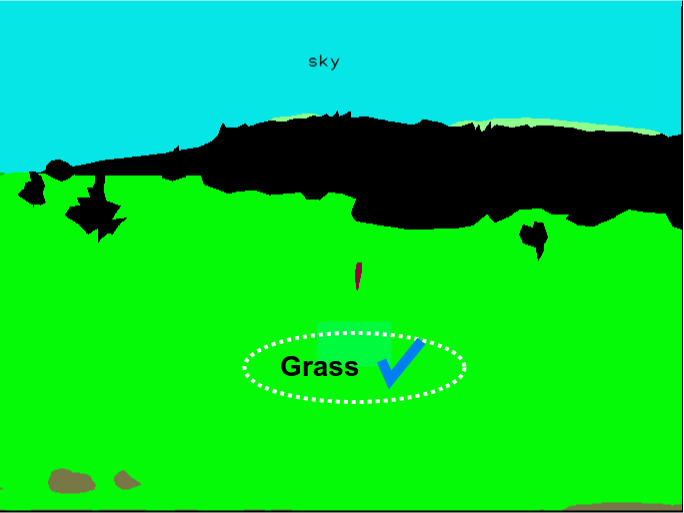} 
        & \includegraphics[valign = c, height=0.06\linewidth, width=0.16\linewidth]{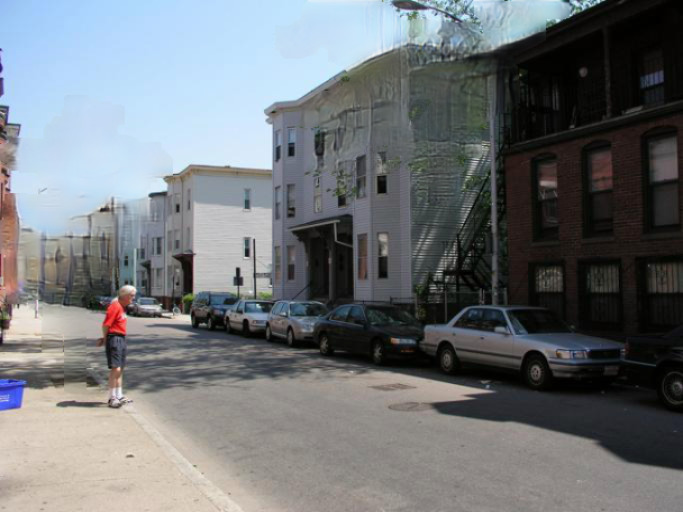}
        & \includegraphics[valign = c, height=0.06\linewidth, width=0.16\linewidth]{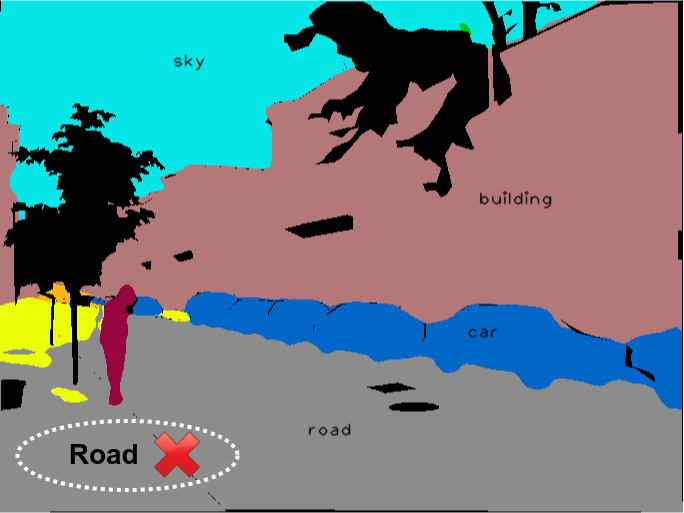}
        & \includegraphics[valign = c, height=0.06\linewidth, width=0.16\linewidth]{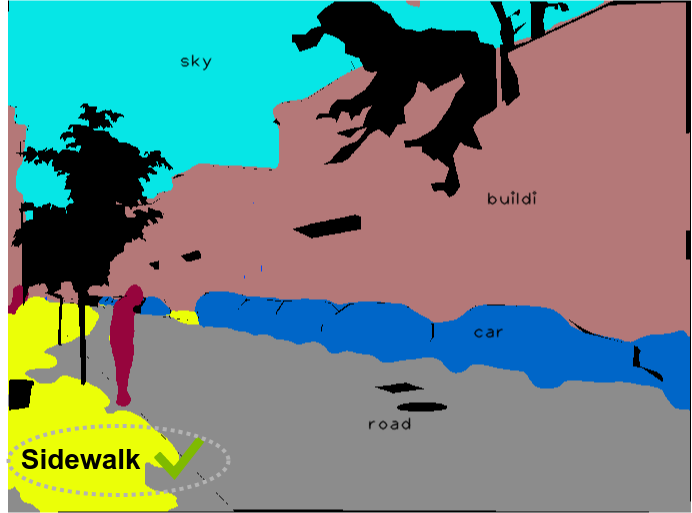}\\
          $\mathcal{I} - tree $ & Upernet  & Ours & $\mathcal{I} - tree$ & Upernet & Ours\\
    \end{tabular}
        }
    \vspace*{-0.5em}
    \caption{Examples of segmentation failures due to removal of a single context object. We see the segmentation of road, sidewalk and grass affected significantly when context objects like signboard, car and tree is removed~(comparing odd and even rows). Model trained with proposed data-augmentation is more robust to these changes.}
    \vspace*{-1em}
    \label{fig:InconsistentSeg}
\end{figure*}

To create edited images with context objects removed, we need a fully automatic object removal model. For this, we utilize ground-truth object masks to remove the desired object and use an in-painting network to fill in the removed region.
We base our in-painting network on the model proposed in~\cite{shetty2018adversarial}, since this inpainter is directly optimized for removal, and can better handle irregular masks~\cite{shetty2018adversarial}, regularly encountered in object removal. More details about the network architecture can be found in the supplementary material.
The above removal method works well for medium sized objects, but struggles for large objects since then the in-painter needs to synthesize most of the image. Hence, we impose size restrictions on the objects we choose to remove to be less than 30\% of the image.
In the classification scenario on the COCO dataset, we consider all 80 object categories for removal.
In the segmentation setting on the ADE20k dataset, we consider only the non-stuff categories (90 categories) for removal and measure the effects of removing these objects on the segmentation of all 140 categories.
The stuff categories include objects like road, sky and field which are typically very large and hard to inpaint and hence are excluded from removal.
An important point to note here is that the in-painter model is not aware of the downstream models and is not optimized to fool or change their decisions. The effects of the in-painter are local and only affects the region the object is removed from.
Qualitative examples in Figures \ref{fig:classInconsist} and \ref{fig:InconsistentSeg} show that the in-painting works reasonably well in the object removal setting.
\subsection{Measuring context dependency}
\label{sec:robust_metrics}
To understand the effect contextual cues have on image-classification and segmentation models, we test them models on edited images where a context object has been removed.
Precisely, given an original image $I$ containing a set of objects $C=\{c_1, c_2 \cdots c_n\}$,  we first create a set of edited images $\mathcal{I}_e = \{I-c_i | c_i \in C\text{ and }\text{removable}(c_i)\}$. Next we test classification and segmentation models on $I$ and $\mathcal{I}_e$ and the check their outputs for consistency with the performed removal as described below for each task. %

\myparagraph{Image-level classification}
Given a trained classifier $S_{c_i}$ for class $c_i$, we will now characterize how robust it is to changes in context of $c_i$. We first obtain classifier scores for the original image $I$, edited image $I-c_i$ with object $c_i$ removed and for the edited set $\mathcal{I}_{\text{owc}} = \{I-c_j: c_j\in I , j\neq i\}$, all of which contain the object $c_i$ but have one context object removed. Ideally, if the classifier $S_{c_i}$ is robust to context changes it should score all the images in $\mathcal{I}_{\text{owc}}$ higher than the image $I-c_i$,
since $I-c_i$ does not contain the object $c_i$ and the images in $\mathcal{I}_{\text{owc}}$ do.
Precisely, a classifier robust to context should satisfy the below in-equality:
\begin{align}
 S_{c_i}(I_{\text{owc}})\geq S_{c_i}(I-c_i), \forall I_{\text{owc}}\in\mathcal{I}_{\text{owc}}
\label{eq:contconst}
\end{align}
\noindent We can count the number of times this condition is violated to quantitatively measure the robustness of the classifier.
{
\thickmuskip=0.4\thickmuskip
\begin{align}
    V^{\text{min}}(c_i) &= \frac{\sum_{I}{\mathbbm{1}\left[\left(\min_{I_{\text{owc}}} {S_{c_i}(I_{\text{owc}})}\right) < S_{c_i}(I-c_i)\right]}}{\sum_{I}{\mathbbm{1}[c_i \in I]}}\\
    V^{\text{mean}}(c_i) &= \frac{\sum_{I}{\mathbbm{1}\left[\mathbb{E}_{I_{\text{owc}}}{\left[S_{c_i}(I_{\text{owc}})\right]} < S_{c_i}(I-c_i)\right]}}{\sum_{I}{\mathbbm{1}[c_i \in I]}}
\end{align}
}
\noindent where $\mathbbm{1}$ is the indicator variable. $V^{\text{min}}(c_i)$ is a strict metric counting instances classifier scores $I-c_i$ higher than any of the edited images, whereas $V^{\text{min}}(c_i)$ is a softer metric counting instances where $I-c_i$ is scored higher than the average score assigned to the edited images.

\myparagraph{Semantic segmentation}
To understand the role context plays in this pixel-level labeling task, we analyze the behaviour of a trained segmentation model by removing one object at a time from the original image. Specifically, we measure how the segmentation correctness of the rest of the image changes~(as compared to segmentation of the original image) when we remove an object from the original image. Given a segmentation model $P$, we compute the intersection-over-union~(IoU) for a class $c_i$ (w.r.t. ground-truth) on the original image $I$ and edited image $I-c_j$. If the IoU value changes more than threshold $\alpha$, we consider the segmentation prediction for class $c_i$ to be affected by removal of $c_j$. Counting these violations we get,
\begin{align}
    AR(c_i,c_j) &= \frac{\sum_I{\mathbbm{1}\left[\left|\Delta\text{IoU}_{c_ic_j}\right|\geq \alpha\right]}}{\sum_I{\mathbbm{1}\left[c_i,c_j \in I\right]}}
    \label{eq:crmetrticSeg}
\end{align}
\noindent where $\Delta\text{IoU}_{c_ic_j}$ is the change in IoU of class $c_i$ with removal of object $c_j$ and $\alpha$ is the change threshold. The matrix $AR(c_i, c_j)$, represents the fraction of images where removing the object $c_j$, affects the segmentation of the object $c_i$ with high values of $AR(c_i, c_j)$ indicating that the segmentation model depends heavily on the presence of the context object $c_j$ to segment $c_i$.
\subsection{Data augmentation with object removal}
\label{sec:da_strategies}
We now present our data augmentation solution to reduce the sensitivity of classification and segmentation models to context distribution. The main idea is to expose these models to training images of object-without-context and context-without-object scenarios. This will help the models deal with the lack of contextual information and hence become more robust to context changes. For this we perform object removal to create edited images with some objects removed and add these edited images to the training batch. Specific details of how to pick objects for removal and how to use them in training for the two tasks are discussed below.

\myparagraph{Classification} For classification we experiment with two strategies to use the edited images in training. In the first approach, we refer to as \emph{Data-aug-rand}, we randomly select one object to remove with uniform probability. Edited image is assigned the same labels as the same as the original image excluding the removed object class. Now we train the classifier with original and data augmented images using simple binary cross-entropy loss. In the second approach referred to as \emph{Data-aug-const} we explicitly optimize for robustness by including the in-equality in \eqref{eq:contconst} in the loss function. To do this, for a randomly selected images in the training batch, we create the full edited image set $\{I-c_i: c_i\in I\}$. Then we can incorporate the robustness constraint as a hinge loss with final loss being weighted sum of cross-entropy loss and the hinge loss $L_h$.
\vspace*{-2mm}
\begin{align*}
    \mathcal{L}_h(I) &= \sum_{c_i\in I}\max{\left[0, S_{c_i}(I-c_i)-\min_{c_j, j\neq i}{S_{c_i}(I-c_j)}\right]} \ \  (5)
\end{align*}\vspace*{-2mm}

\myparagraph{Segmentation} We also perform data augmentation on the segmentation task by creating edited images by selectively removing objects. The edited images can be used in training the segmentation model in two ways. First we can simply ignore the removed pixels and train the model to predict the original ground-truth labels on the rest of the image~(\emph{Ignore}). This helps the model learn that the labeling of a pixel should not be affected by the removal of a context object. Alternatively, we can explicitly tell the model that the removed object is not present by minimizing the likelihood assigned to the removed class at the edited pixel locations~(\emph{Negative loss}). %
The next question is how to sample the objects to remove. We explore three different strategies to select these objects. The first strategy, \emph{Random},  selects one random object to remove from the objects present in the image with uniform probability. However, sometimes the \emph{Random} strategy can select very large object for removal, which can harm the quality of the edited image. To address this we use the \emph{Sizebased} strategy, which select objects based on their relative sizes in the image, giving higher probability to selecting smaller objects. The probability for picking an object is computed as
    $p(c_i,I) \propto \left[\frac{\sum_{c_i \in I}{a(I,c_i)}}{a(I,c_i)}\right]$
where $a(I,c_i)$ is the area of the class $c_i$ in image $I$.%
We also explore a hard negative mining based strategy, where we create harder training examples for the segmentation model by removing easy classes.  This allows the model to focus on segmenting the harder classes while also becoming robust to context. Concretely, in this \emph{HardNegative} strategy we monitor the average cross-entropy segmentation loss $l_{\text{avg}}(c_i)$ for an object class $c_i$ and calculate the probability of removal of $c_i$  as inversely proportional to $l_{\text{avg}}(c_i)$.%

%% file: results.tex
This section presents the results of our analysis of how much the contextual information influences the performance of image classification and segmentation models. Using the robustness metrics defined in \secref{sec:robust_metrics}, we discover that the classification predictions on many well-performing classes are sensitive to context, and perform poorly on object-without-context and context-without-object images. Similar results are also found in the segmentation setting with the model depending heavily on context objects to correctly segment classes like \emph{road}, \emph{sidewalk}, \emph{grass}. We also present results from our data-augmentation strategies, which help reduce this context dependence and improve robustness, without sacrificing performance.

\subsection{Image level classification}
\subsubsection{Experimental setup for classification}
\vspace*{-1mm}
\myparagraph{Training data} We run our classification experiments on the COCO dataset~\cite{lin2014microsoft}, which contains 80 labeled object classes in their natural contexts. The dataset also has bounding box and segmentation annotation for each object. We use image-level labels to train the classifiers and use the object segmentation masks to test them with object removal.

\myparagraph{Out-of-context testing} Apart from testing the classifier models on regular COCO data we conduct additional experiments to quantify the performance in out-of-context scenarios with natural images. We divide the COCO images into two splits: the first split \emph{Co-occur} with images having at least two objects in them and the second split \emph{Single} with images containing a single object. The \emph{Full} split is all images combining \emph{Co-occur} and \emph{Single}. The idea behind this splitting of the dataset is to separate out images where objects occur in their context~(\emph{Co-occur}) and images where object occur alone without the usual co-occurring context objects~\emph{Single}. Now we can train our models on the \emph{Co-occur} split and test it on the \emph{Single} split to measure, using only real images, how a classifier trained with only co-occurring objects performs when objects appear without the context seen in training.
Additionally we also test our COCO trained models on the \emph{UnRel} dataset~\cite{Peyre17} which contains natural images with objects occurring in unusual contexts and relationships. We keep the classes which map to one of the 80 object classes in COCO, leaving 29 classes and 1071 images in the UnRel dataset.

\myparagraph{Baseline classifier} The image-level classification model we test is based on the architecture proposed in~\cite{oquab2015object}. It consists of a Imagenet~\cite{imagenet_cvpr09} pre-trained VGG-19 network for feature extraction network followed by two convolution layers, global max-pooling layer and a linear classification layer with sigmoid activations. The model is trained with binary cross-entropy loss. We train and test the model at single scale at 256x256 resolution, to simplify the analysis. Our classifier achieves similar mAP on real coco data as reported in \cite{oquab2015object}, with our mAP slightly lower (0.600 vs 0.628 in \cite{oquab2015object}) due to single scale training and testing. %

\begin{figure*}[t]
    \adjustbox{valign=t}{
    \begin{minipage}[t]{.58\linewidth}
        \centering
      \begin{minipage}[b]{1.0\linewidth}
        \newcommand*{\mrowt}[1]{\multirow{2}{*}{#1}}
        \footnotesize
        \setlength{\tabcolsep}{0.1mm}
        \begin{tabular}{x{2cm}c ccc cc c}
        \toprule
        \multirow{2}{*}{Model} & \multirow{2}{*}{\parbox{1.8cm}{\centering Training Data}}& \multicolumn{3}{c}{COCO test set}&  \multicolumn{2}{c}{Robustness Metrics}& UnRel\\\cmidrule(lr){3-5}\cmidrule(lr){6-7}
                               & & Full $\uparrow$& Co-occur $\uparrow$& Single $\uparrow$& $V^{\text{min}} \downarrow$& $V^{\text{mean}} \downarrow$ &dataset $\uparrow$ \\\cmidrule(lr){1-1}\cmidrule(lr){2-2}\cmidrule(lr){3-5}\cmidrule(lr){6-7}\cmidrule(lr){8-8}
            Baseline        & Full~(39k) & 0.60    & 0.57    & 0.62    & 34\% & 24\%        & 0.50    \\
            Data-aug-rand   & Full~(39k) & \bf0.61 &\bf0.58  & \bf0.65 & 32\% & 22\%        & \bf0.54 \\
            Data-aug-const  & Full~(39k) & 0.60    &\bf0.58  & 0.63    & \bf 25\% & \bf14\% & 0.52     \\\cmidrule(lr){1-1}\cmidrule(lr){2-2}\cmidrule(lr){3-5}\cmidrule(lr){6-7}\cmidrule(lr){8-8}
            Baseline        & Co-occur~(30k)& 0.56    & 0.55    & 0.58    & 34\% & 24\%       & 0.46      \\ 
            Data-aug-rand   & Co-occur~(30k)& \bf0.58 & \bf0.57 & \bf0.60 & 31\% & 21\%       &0.49       \\
            Data-aug-const  & Co-occur~(30k)& \bf0.58 & \bf0.57 & \bf0.60 & \bf27\% & \bf15\% & \bf0.51   \\
         \bottomrule
        \end{tabular}
        \vspace*{-2mm}
        \captionof{table}{ Effect of data augmentation on classification model}
        \label{tab:da_classification}
    \end{minipage}\\[0.5em]
\begin{minipage}[b]{1.0\linewidth}
        \footnotesize
    \setlength{\tabcolsep}{8pt}
    \begin{tabular}{p{10mm} p{8mm}p{8mm} p{8mm}p{8mm} p{8mm}p{8mm}}
    \toprule
      \multirow{2}{*}{Model} & \multicolumn{2}{c}{all (407 images)}& \multicolumn{2}{c}{with car (258)}& \multicolumn{2}{c}{without car (149)}\\
      & Road & Sidewalk &Road & Sidewalk &Road & Sidewalk \\\cmidrule(lr){1-1}\cmidrule(lr){2-3}\cmidrule(lr){4-5}\cmidrule(lr){6-7}
      Upernet   &  0.81 & 0.59 & \bf0.86 & \bf 0.67 & 0.68 & 0.40 \\
      DataAug  &  \bf0.82 & \bf0.60 & \bf0.86 & 0.65 & \bf0.72 & \bf0.46 \\
      \bottomrule
    \end{tabular}
     \vspace*{-2mm}
    \captionof{table}{Comparing the performance of road and sidewalk\\ segmentation on natural images with and without cars.}
    \label{tab:daseg_road}
\end{minipage}
\end{minipage}
}\hfill
\adjustbox{valign=t}{
    \begin{minipage}{.42\linewidth}
        \includegraphics[width=0.8\linewidth]{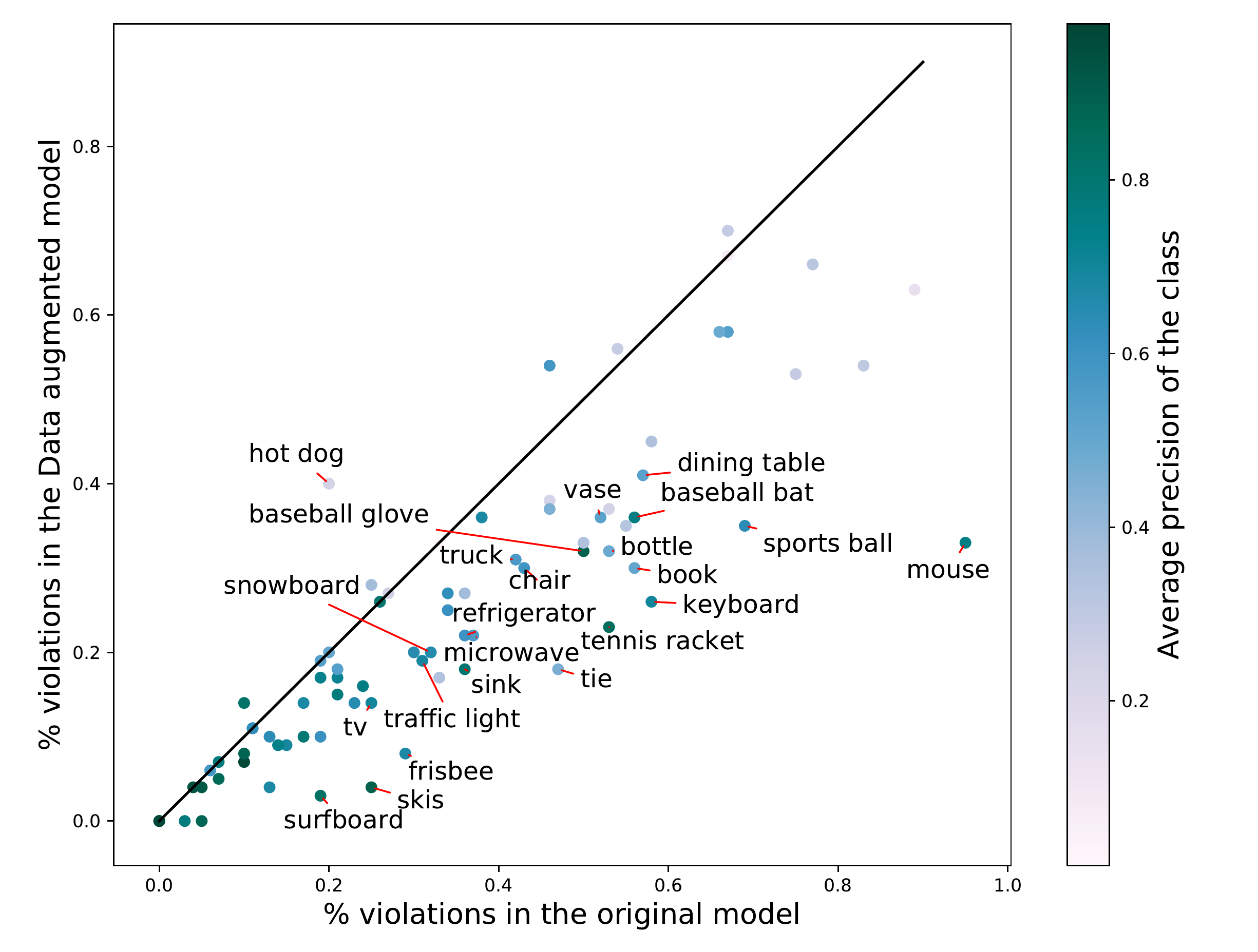}
        \captionof{figure}{Comparing the \% of violations in different classes with and without data augmentation. Points below the diagonal line show improvement with data-augmentation and the ones above degrade. The colors denote the average precision.}
\label{fig:PerclassViol_DA_comparison}
\end{minipage}}
\end{figure*}
\subsubsection{Analyzing classifier robustness to context}
To measure the robustness of the trained classifier to context, we test it on real images and edited images and compute the robustness scores $V^{\text{min}}$ and $V^{\text{mean}}$ as described in \secref{sec:robust_metrics}.
\Tableref{tab:da_classification} shows the robustness scores averaged over all classes computed on the COCO test along with the standard performance metric mean average precision (mAP) for the baseline classifier~(first row). We can see that, despite achieving good mAP (0.6), the baseline classifier trained on full data performs poorly in-terms of robustness metrics. In about 34\% of cases the model violates the context consistency requirement of \eqref{eq:contconst}. This means in 34\% cases, the classifier scores images without the target object higher than an image where object is present but a context object has been removed. Comparing the per-class robustness score, $V^{\text{min}}(c_i)$ and the per-class average precision~(AP)~(see supplementary for visualization), we see that good performance in AP does not mean the classifier is robust to context. Many classes like mouse, keyboard, sink, tennis racket etc, which are performing well in AP ($\geq 0.8$), but have poor robustness to changes in context ($V_o^{\text{min}} \geq 50\%$). In extreme case, the \emph{mouse} classifier violates the consistency in more than 90\% of cases, despite having very good AP (0.88). This indicates that the classifiers are relying too much on contextual evidence to detect the objects but perform poorly when tested on images where the context distribution is different from  training.%

We visualize these violations in \figref{fig:classInconsist}. In the first row we see the case where the keyboard classifier scores the image with the keyboard removed higher~(4.67) than the image with the keyboard but with the monitors removed (1.99). Similarly we see the skateboard and the frisbee classifiers relying on person to hallucinate the respective objects. The violations shown in the first three rows of \figref{fig:classInconsist} occur in objects with high co-occurrence dependence with other classes. 
However, such context violations can also be seen in classes like \emph{person} which occur in very different contexts as seen in the last row of \figref{fig:classInconsist}. Here, the violation occurs in a difficult image where the \emph{person} is small, but a more distinct class with co-occurrence dependence on person is clearly visible (\emph{kite}). The classifier seems uses the \emph{kite} context class to hallucinate that there is a \emph{person}, even when the \emph{person} has been removed.

\subsubsection{Data augmentation to improve robustness}
We train two variants of the data-augmented image classification models as described in \secref{sec:da_strategies}. The first \emph{Data-aug-rand} learns with standard cross-entropy loss on the edited images with a random object removed and the second \emph{Data-aug-const} which is optimized directly for robustness using a set of edited images and hinge loss.

\myparagraph{Quantitative results} We present the evaluation of the data-augmented and the baseline models in \tableref{tab:da_classification}. On models trained with \emph{Full} training data, the data-augmented model \emph{Data-aug-rand} provides a small improvement in overall mAP on the COCO test set (0.61 vs 0.60). However measuring the performance on the two splits \emph{Co-occur} and \emph{Single} reveals that the improvement is significant on the \emph{Single} split (0.65 vs 0.62), indicating that the data augmentation helps the classifier better deal with out of context objects.
This is also seen when comparing the performance of the two models on the UnRel dataset, where \emph{data-aug-rand} significantly improves over the baseline model (0.54 vs 0.50).
This improved robustness of the data augmented classifier to context changes is also measured by our robustness metrics $V^{\text{min}}$ and $V^{\text{mean}}$.
\emph{Data-aug-rand} classifier makes overall 2\% less violations under both worst-case~($V^{\text{min}}$) and average-case~($V^{\text{mean}}$) context changes.
Directly optimizing the robustness constraints allows the model \emph{Data-aug-const} to significantly improve upon the baseline model in robustness metrics, while still obtaining improvement in the performance metrics. It exhibits much less worst-case~(25\% vs 34\% for baseline) and average-case violations~(14\% vs 24\% for baseline), while improving the performance in the UnRel dataset~(0.52 mAP vs 0.50 for baseline). 
The benefit of optimizing for robustness is clearly seen when we constrain the training data to the \emph{Co-occur} set, where the classifier never sees objects alone. Baseline model trained on the \emph{Co-occur} set drops in performance on the \emph{Single}~(0.58 from 0.62 on when trained on \emph{Full}) and the UnRel test sets~(0.46 vs 0.50 with \emph{Full}) . However, with data augmentation and enforcing robustness constraints, we can recover some of this performance. On the \emph{Single} test set \emph{Data-aug-const} model trained on \emph{Co-occur} set gets 0.58 mAP compared to 0.60 by baseline model trained on full data and even surpass it on the UnRel test set with 0.51 mAP. This shows that the data augmented model is able to overcome the contextual bias in the training set and perform well in unseen contexts.

When we compare the per-class robustness metrics between regular and data augmented models~(data-aug-const), as shown in the \figref{fig:PerclassViol_DA_comparison}, we see that data-augmentation significantly reduces the worst case violations~($V^{\text{min}}$) on well-performing classes. For example, $V^{\text{min}}$ drops from 95\% to less 36\% for the mouse class and from 58\% to 28\% for the keyboard class. %
The effect of this increased robustness is seen in qualitative examples in \figref{fig:classInconsist}. 
For example, in the first row the baseline keyboard classifier gives too much weight to evidence from \emph{monitor} and scores the image with only \emph{monitor} higher than the only keyboard. However, the data augmented model correctly orders the images. 
\subsection{Semantic segmentation}
So far, we have seen that multi-label classification models suffer from sensitivity to context, with classifiers often mixing up contextual and visual evidence. Next we will measure the context sensitivity of models in a more local and strongly supervised task of semantic segmentation.

\subsubsection{Experimental setup for segmentation}
\myparagraph{Training and test data} We conduct our semantic segmentation experiments primarily on the ADE20k dataset~\cite{zhou2017scene} containing 140 categories of labeled objects, in different settings. Some of the 140 classes are typical background classes like \emph{sky}, \emph{sea} and \emph{wall} and are large and difficult to in-paint and are hence excluded from removal. 

\myparagraph{Out-of-context testing} Following the process in image-level classification, we also measure the performance of the segmentation models on real out-of-context data. This in done in two ways. First, we train the segmentation model in a restricted setting with only three classes \emph{car}, \emph{road} and \emph{sidewalk}. Now, we can again make two splits of the training and testing images into the \emph{Co-occur} split of images with at-least two objects~(3317 images) and the \emph{single} split with only a single object~(1693 images). Then we train the segmentation models on \emph{co-occur} split and test on \emph{single} split to see how well it can perform  segmentation without context. 
Additionally we also test the models trained with ADE20k data on the Pascal-context dataset~\cite{mottaghi2014role} in order to measure the performance under a different context distribution.  This is done by manually mapping the 59 labels in the pascal-context to ADE20k labels and restricting the segmentation model to produce only the mapped labels.

\myparagraph{Baseline segmentation model}
We use the recent UperNet~\cite{xiao2018unified} model, with good results on the ADE20k, as our baseline segmentation model. We train the variant with the Resnet-50 encoder and a Upernet decoder with batch size of 6 images~(maximum that fit in GPU) and with the default hyper-parameters suggested by the authors. This model achieves mean intersection-over-union~(mIoU) of 0.377 and accuracy of 78.19\% with single scale testing. %

\subsubsection{Context in semantic segmentation}
We analyze robustness of the segmentation models to context by removing objects and computing the matrix $AR(c_i, c_j)$ presented in \secref{sec:robust_metrics}, which measures the \% of images where removal of object $c_j$ significantly affects segmentation of object $c_i$. 
The matrix $AR(c_i,c_j)$ we obtain for the Upernet model in ADE20k dataset is a sparse matrix with sharp peaks~(see supplementary for a visualization). This indicates that the classes depend on specific context objects and are significantly affected by their removal. The sparsity also indicates that the effects on the segmentation are due the class being removed and not in-painting artifacts~(otherwise the segmentation would be affected by all removal). Some of dependencies we discover in  $AR(c_i,c_j)$ are reasonable and harmless, for example between \emph{pot} and \emph{plant}~($AR=50\%$). Once you remove the \emph{plant}, \emph{pot} looks more like a \emph{trash can} and the segmentation model often flips the label to \emph{trash can}. However other dependencies are spurious and not desirable. For example, we notice that often the segmentation model uses presence of \emph{car} to differentiate between \emph{road} and \emph{sidewalk}. Removing \emph{car} affects the IoU of the \emph{road} and \emph{sidewalk} in 21\% and 22\% of cases respectively. This dependence is undesirable, and can be catastrophic in applications like self-driving cars.

We show qualitative examples where removal affects segmentation of Upernet model in \figref{fig:InconsistentSeg}. The first two rows show the cases where removal of an object negatively impacts the segmentation of other objects. This include cases where removal of \emph{street sign} and \emph{car} severely affects segmentation of \emph{road} and \emph{sidewalk}, and a case where removal of \emph{trees} affects segmentation of \emph{grass}. We can see from these examples that while edit on the image is small and local, the effects of this removal on segmentation prediction is not local. Removal of a small objects can have drastic effects on segmentation in a far-away region. 

\subsubsection{Data augmentation for segmentation}
\begin{table}[t]
    \newcommand*{\mrowt}[1]{\multirow{2}{*}{#1}}
    \footnotesize
    \setlength{\tabcolsep}{1mm}
    \centering
    \begin{tabular}{lc cc}
    \toprule
    \multirow{2}{*}{Model} & Removed &  \multicolumn{2}{c}{ADE20k}\\\cmidrule(lr){3-4}
              & Pixels & mIoU & Acc\\\cmidrule(lr){1-2}\cmidrule(lr){3-4}
        Upernet\cite{xiao2018unified}      & - & 0.377 & 78.31 \\\cmidrule(lr){1-2}\cmidrule{3-4}
        DA (random)        & Ignore   & 0.320 & 75.2 \\
        DA (sizebased)     & Ignore   & 0.379 & 78.31 \\
        DA (hard negative) & Ignore   & 0.375 & 77.8 \\\cmidrule(lr){1-2}\cmidrule{3-4}
        DA (sizebased)     & Negative & 0.377 & 78.25 \\
        DA (hard negative) & Negative & \bf0.385& \bf78.47\\
     \bottomrule
    \end{tabular}
    \caption{Data augmentation results on ADE20k dataset}
    \label{tab:da_segmentation}
\end{table}
Next we will look at the results of using data-augmentation for segmentation models. For this purpose we train the Upernet~\cite{xiao2018unified} based data-augmented models on the ADE-20k dataset with on three different strategies for selecting the object to remove as discussed in \secref{sec:da_strategies}. 

\myparagraph{Quantitative results} \Tableref{tab:da_segmentation}, shows the results comparing the data-augmented models with the baseline Upernet model. We can see that random sampling strategy, which worked well in image classification, fails here leading to drop in performance. This is because, many object categories in ADE20k dataset are large and difficult to remove like bed, sofa and mountain and random strategy suffers by picking these. Instead when we switch to size-based and hard-negative based sampling, we see that the performance improves and the the size-based sampling model achieves the best mIoU of the three models (0.379). Applying negative likelihood loss on the removed object class gets further improvement when combined with hard negative sampling. This model also improves upon the Upernet baseline~(achieving 0.385 IoU vs 0.377 by Upernet), despite the fact that the removal based data-augmentation is designed to make the model more robust to contextual variations.%

\begin{figure}
\centering
    \includegraphics[width=0.65\linewidth]{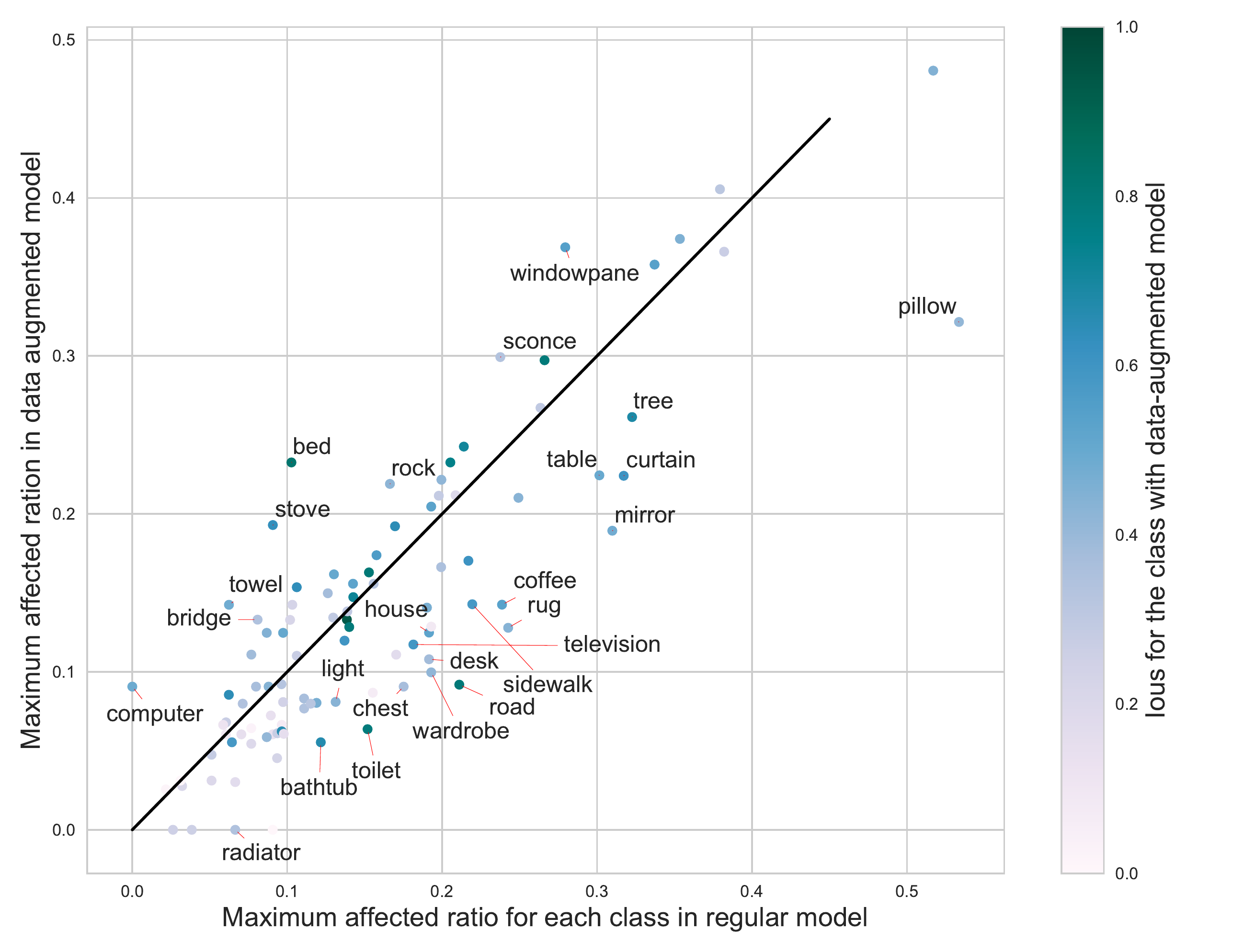}
    \caption{Comparing the context sensitivity of different classes with and without data augmentation with $\max_{c_j}{AR(c_i, c_j)}$ metric. Points below the diagonal improve with data-augmentation. The color denotes the mIoU. }%
\label{fig:PerclassViolSegm_DA_comparison}
\end{figure}

To understand how data-augmentation impacts sensitivity to context, \Figref{fig:PerclassViolSegm_DA_comparison} visualizes the maximum sensitivity of a class to removal of other classes, $\max_{c_j}{AR(c_i,c_j)}$ for different classes with and without data-augmentation. We see that for majority of classes robustness to context improves with data augmentation. For example \emph{pillow} class is only affected 32\% of the time with context changes, compared to 53\% before data augmentaion. Similary, \emph{road} and \emph{sidewalk} classes are only affected 9\% and 14\% of the time respectively, compared to 21\% and 22\% before. 
This improved robustness translates into better generalization to real out-of-context data. We can see this in~\tableref{tab:daseg_road} where the performance of the \emph{road} and \emph{sidewalk} segmentation is measured on the validation set on images with and without cars. On the full set and on the split with cars, we see that the performance of the baseline Upernet and our augmented model (DA hard negative with negative loss) is equivalent. However, when we look at only images without car, the Upernet model performs significantly worse in both road ~(0.68 vs 0.72 for ours) and sidewalk~(0.40 vs 0.46 for ours) segmentation. This quantitatively shows that the baseline model struggles to distinguish between \emph{road} and \emph{sidewalk} without \emph{car} in the image, whereas our data augmentation is more robust and performs well even without context~(car).

We also see the benefit of data augmentation in experiments on restricted \emph{Co-occur} training set and on the Pascal-context dataset. 
Our data augmented model outperforms the Upernet model~(both trained on the ADE20k dataset) when tested on the Pascal-context dataset in both mIoU and pixel accuracy. While the Upernet model achieves mIoU of 0.284 and pixel accuracy of 61.3\% our data augmented model achieves 0.293 and 62.10\% respectively, indicating that it is able to generalize better when tested on a dataset with different context distribution than one seen during training.
\Tableref{tab:da_seg_rest} presents the experiments with the \emph{Co-occur} training set in the three class setting. First we can see that when we switch from training on \emph{Full} training data to \emph{Co-occur} split (containing only images with atleast two objects), the performance of the Upernet greatly drops on the \emph{Single} test split~(from 0.67 to 0.52). This is indicates that the model overfits to the context it sees, and is not able to segment objects when it seeing them out of context. However, with data-augmentation we generate images of objects without context, and can recover most of this performance loss~(0.646). Surprisingly, data-augmented model trained on smaller \emph{co-occur} data also outperforms the baseline trained with \emph{Full} data when tested on the \emph{co-occur} split. %

Qualitative examples in \figref{fig:InconsistentSeg} also show the effect of increased robustness to context. While the baseline Upernet model is affected by context object removal causing drastic changes in predictions of other regions, our data augmented model is more stable. For example the removal of \emph{signboard},  \emph{car} or \emph{tree} does not effect the segmentation of the \emph{road} or \emph{sidewalk} by our model. %

\begin{table}[t]
    \newcommand*{\mrowt}[1]{\multirow{2}{*}{#1}}
    \footnotesize
    \setlength{\tabcolsep}{1mm}
    \centering
    \begin{tabular}{lc ccc }
    \toprule
    Model & Training Data& Full & Only Cooccur & Only Single \\\cmidrule(lr){1-1}\cmidrule(lr){2-2}\cmidrule(lr){3-5}\cmidrule(lr){3-5}
        Upernet  & Full~(5k) & 0.774 & 0.797 & 0.670 \\
        Data Aug & Full~(5k) & 0.742 & 0.754 & \bf0.675 \\\cmidrule(lr){1-5}
        Upernet  & Co-occur~(3.3k) & 0.680 & 0.713 & 0.520 \\ 
        Data Aug & Co-occur~(3.3k) & \bf0.82 & \bf0.86 & 0.646 \\
     \bottomrule
    \end{tabular}
\caption{Experiments in three class setting on ADE20k}%
    \label{tab:da_seg_rest}
\end{table}

%% file: conclusions.tex
We have presented a methodology to analyze and quantify the context sensitivity of image classification and segmentation models, based on editing images to remove objects and measuring the effect on the target model output.  Our analysis shows that despite good performance in-terms on mAP, classifiers for certain classes like keyboard, mouse, skateboard are very sensitive to context objects and perform poorly when seen out of context. In semantic segmentation setting, our analysis shows similar dependency between classes. For example we discover that the model depends on the  presence of car to segment roads and sidewalk and fails drastically when the car is not present in the image. We present a data augmentation scheme based on object removal to mitigate this and make the classification and segmentation models more robust to context changes. Our experiments show that the proposed data augmentation scheme can help models generalize to out of context scenarios without losing performance in standard setting, indicating that the data augmented models better balance contextual and visual information.

%% file: supplementary_material.tex
\section{Object removal model}
\label{sec:remmodel}
To remove objects we use the ground truth segmentation masks and dilate them by a small factor (5 in the coco dataset and 7 in the ADE20k dataset). This dilated mask is multiplied with the input image to remove the target object from the image. Then the masked image and the mask is passed to an in-painting network which fills the masked area with a plausible background texture. We use the in-painting architecture and the training procedure proposed in~\cite{shetty2018adversarial}. \Tableref{tab:removalModel} and \ref{tab:removalModelResidual} present the detailed architecture of the in-painting network. We will make the code and pre-trained in-painter models available after the review process.

\section{Analyzing robustness to context}
\label{sec:robus}
In the main paper, we presented our analysis showing that the classification and segmentation models are sensitive to context and their predictions are significantly affected when presented with edited images with context objects removed. In the following sub-sections we present additional visualizations to support these arguments.

\begin{table}[t]
    \footnotesize
    \setlength{\tabcolsep}{1mm}
    \centering
    \begin{tabular}{c}
    \toprule
    Masked Image + mask\\
    Conv 4x4, 64 filters, stride 1\\ 
    Conv 4x4, 128 filters, stride 2\\ 
    Conv 4x4, 256 filters, stride 2\\ 
    Conv 4x4, 512 filters, stride 2\\ 
    Residual Block, 256 filters\\ 
    Residual Block, 256 filters\\ 
    Residual Block, 256 filters\\ 
    Residual Block, 256 filters\\ 
    Residual Block, 256 filters\\ 
    Residual Block, 256 filters\\ 
    Upsample + Conv 3x3, 256 filters\\ 
    Upsample + Conv 3x3, 128 filters\\ 
    Upsample + Conv 3x3, 64 filters\\ 
    Conv 7x7, 3 filters, stride 1\\ 
    \bottomrule
    \end{tabular}
    \caption{In-painting model architecture starting with input in the first row to the output layer in the last. Each convolutional layer  is followed by a Instance Norm layer and a Leaky Relu non-linearity with slope 0.1}
    \label{tab:removalModel}
\end{table}

\begin{table}[t]
    \footnotesize
    \setlength{\tabcolsep}{1mm}
    \centering
    \begin{tabular}{c}
    \toprule
    Conv 3x3, n filters, stride 1\\ 
    Instance Norm\\ 
    Leaky Relu (slope 0.1)\\ 
    Conv 3x3, n filters, stride 1\\ 
    Instance Norm\\ 
    Leaky Relu (slope 0.1)\\ 
    \bottomrule
    \end{tabular}
    \caption{Architecture of the residual block with n filters}
    \label{tab:removalModelResidual}
\end{table}

\subsection{Image-level Classification}
\label{sec:robusClas}
\begin{figure}
\centering
    \includegraphics[width=1.0\linewidth]{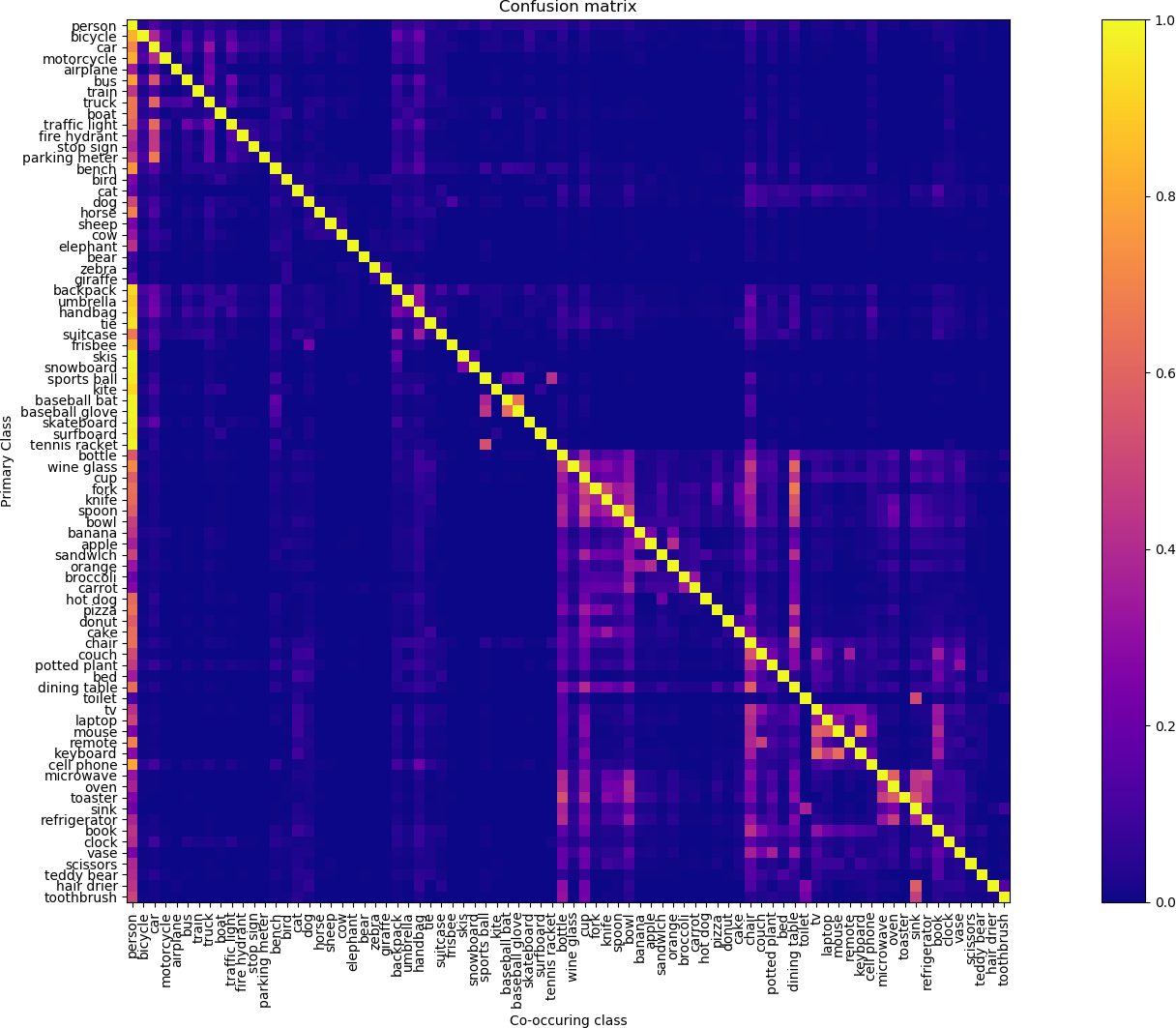}
    \caption{Co-occurrence ratio, $N(c_i, c_j)$ of objects on the COCO dataset}
\label{fig:cococooccur}
\end{figure}

\myparagraph{Co-occurrence of objects} An important factor which causes the image-level classification models to use contextual dependencies is the co-occurrence distribution of objects.Many objects in COCO have a strong co-occurrence relation with other objects.
We quantify this using the normalized co-occurrence counts for each object with others given by $$NC(c_i,c_j) = \frac{\text{Count}(c_i\cap c_j)}{\text{Count}(c_i)}$$. This matrix is visualized in \figref{fig:cococooccur}. $NC(c_i,c_j)$ takes value between 0 and 1 and represents the fraction of images containing object $c_i$, which also contains object $c_j$. We can see that, for classes like skateboard, surfboard, tennis racket and handbag, this ratio is very high ($\geq90\%$) with the person class, since these classes often occur with a person holding or riding them. We also see that the matrix is not symmetric. This is because, while the skateboard might occur always with a person, but person class occurs in various contexts without skateboard. However, for some groups of objects like mouse, keyboard and monitor, and spoon, fork, and cup have symmetric co-occurrence relationship. 

For many categories, including the cases discussed above, the co-occurrence ratio is very high ($>$ 60\%). This causes problems for object classifiers of these categories, as we see in the analysis presented in the main paper. When a small or difficult to detect object class like mouse or skateboard, frequently co-occurs with a more easy to detect object class like monitor or person, the classifiers tend to overuse the contextual relationship for making their classification decisions instead of visual evidence for the object of interest. This leads to failures when the context is different or the object occurs without context. %

\begin{figure}
\centering
    \includegraphics[width=1.0\linewidth]{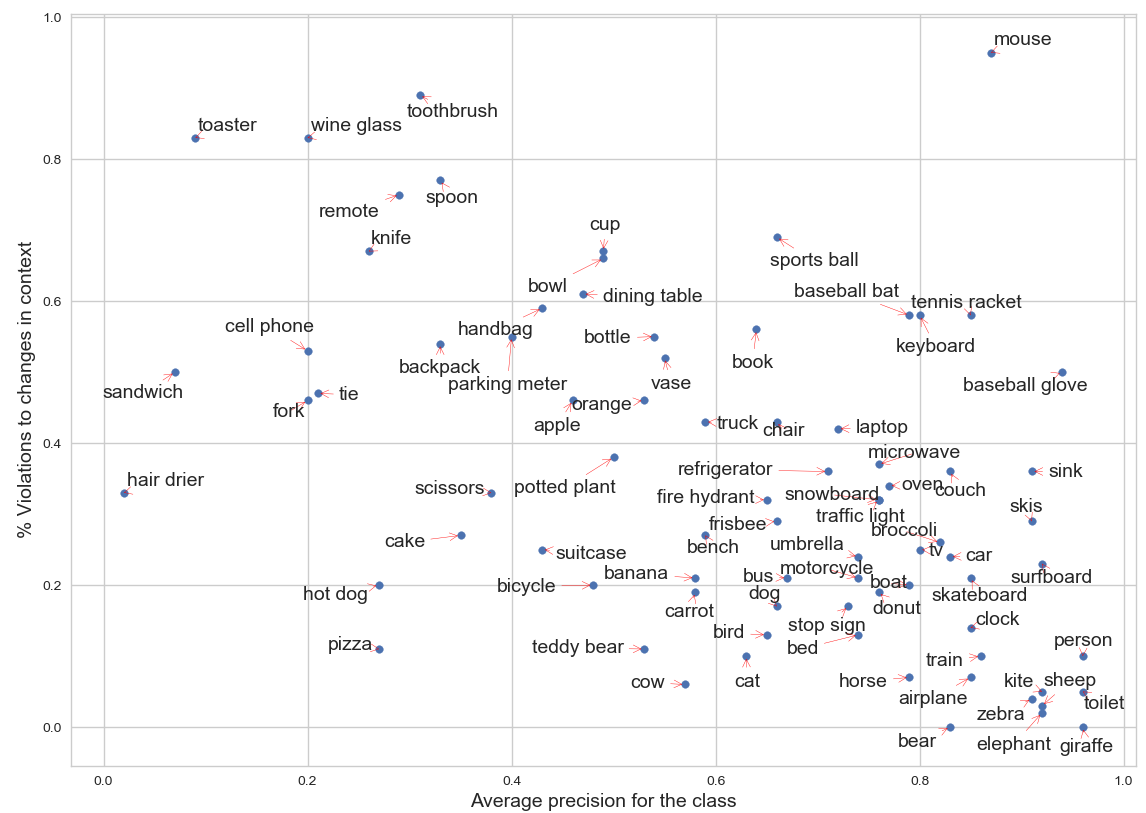}
    \caption{Comparing class-wise average precision to the \% of violations in changes to context. Many well-performing categories (high mAP), have high percentage of violations, including mouse, tennis racket, keyboard, book, and sink.}
\label{fig:PerclassViol}
\end{figure}

\begin{figure}
    \setlength{\tabcolsep}{0.1mm}
    \noindent\makebox[\linewidth]{
    \scriptsize
    \begin{tabular}{x{1.8cm}cx{5mm}c}
        & \bf\small \textcolor{blue}{Object} without \textcolor{magenta}{Context} && \bf\small \textcolor{magenta}{Context} without \textcolor{blue}{Object}\\[0.5em]
         \includegraphics[valign =c, width=0.9\linewidth]{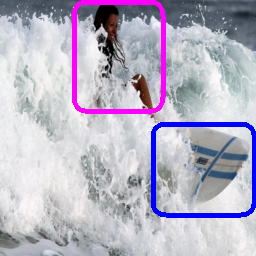} Original
        & \includegraphics[valign = c, width=0.32\linewidth]{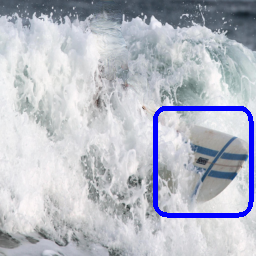}&
        & \includegraphics[valign = c, width=0.32\linewidth]{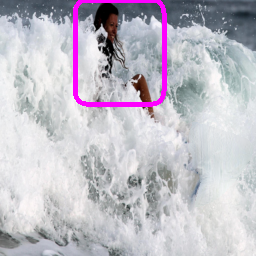}\\
         \bf Regular       & $S(\text{\it surfboard})=\textcolor{red}{2.60}$\textcolor{red}{\Lightning}         &\multirow{2}{*}{\large$\boldmath\geq$ }&$S(\text{\it surfboard)} = \textcolor{red}{3.78}$\textcolor{red}{\Lightning}\\
         \bf Ours & $S(\text{\it surfboard})=\textcolor{PineGreen}{1.81}$    &&$S(\text{\it surfboard}) = \textcolor{PineGreen}{-1.80}$\\[0.5em]

         \includegraphics[valign = c, width=0.9\linewidth]{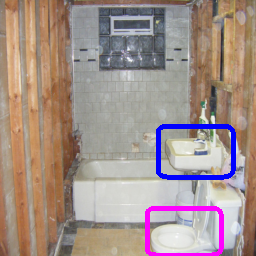} Original
        & \includegraphics[valign = c, width=0.32\linewidth]{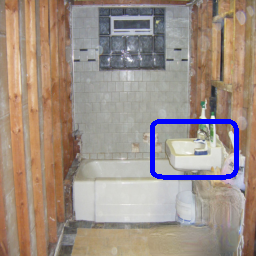}&
        & \includegraphics[valign = c, width=0.32\linewidth]{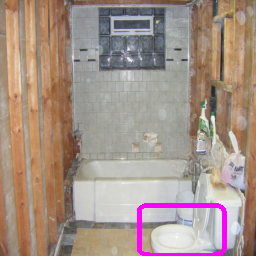}\\

         \bf Regular       & $S(\text{\it sink}) = \textcolor{red}{-0.74}$\textcolor{red}{\Lightning}       &\multirow{2}{*}{\large$\boldmath\geq$ }& $S(\text{\it sink}) = \textcolor{red}{0.60}$\textcolor{red}{\Lightning}         \\
         \bf Ours & $S(\text{\it sink}) = \textcolor{PineGreen}{-0.01}$ && $S(\text{\it sink}) = \textcolor{PineGreen}{-0.24}$ \\
 
         \includegraphics[valign = c, width=0.9\linewidth]{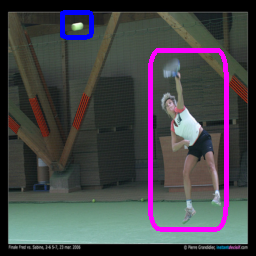} Original
        & \includegraphics[valign = c, width=0.32\linewidth]{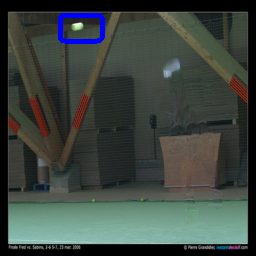}&
        & \includegraphics[valign = c, width=0.32\linewidth]{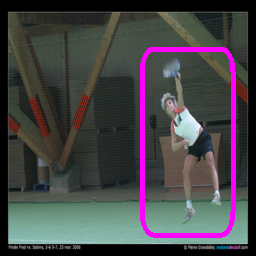}\\
         \bf Regular       & $S(\text{\it ball})=\textcolor{red}{1.96}$\textcolor{red}{\Lightning}    &\multirow{2}{*}{\large$\boldmath\geq$ }& $S(\text{\it ball}) = \textcolor{red}{2.50}$\textcolor{red}{\Lightning}\\
         \bf Ours & $S(\text{\it ball})=\textcolor{PineGreen}{4.85}$    && $S(\text{\it ball}) = \textcolor{PineGreen}{-1.24}$ \\[0.5em]

          \includegraphics[valign = c, width=0.9\linewidth]{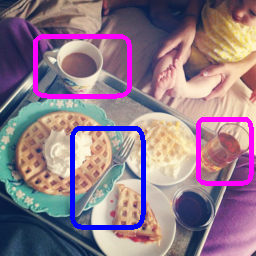} Original
        & \includegraphics[valign = c, width=0.32\linewidth]{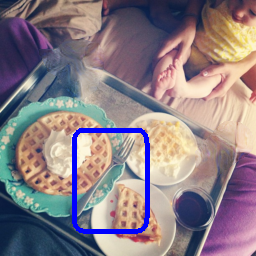}&
        & \includegraphics[valign = c, width=0.32\linewidth]{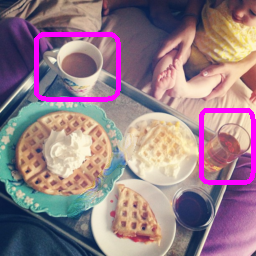}\\

         \bf Regular       & $S(\text{\it fork}) = \textcolor{red}{-2.12}$\textcolor{red}{\Lightning}      &\multirow{2}{*}{\large$\boldmath\geq$ }& $S(\text{\it fork}) = \textcolor{red}{-1.22}$\textcolor{red}{\Lightning}      \\ 
         \bf Ours & $S(\text{\it fork}) = \textcolor{PineGreen}{-3.98}$&& $S(\text{\it fork}) = \textcolor{PineGreen}{-5.06}$\\[0.5em]

          \includegraphics[valign = c, width=0.9\linewidth]{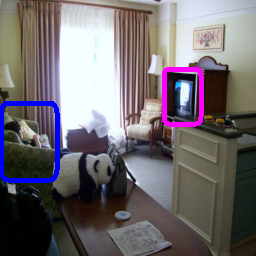} Original
        & \includegraphics[valign = c, width=0.32\linewidth]{classification/correlErrors/COCO_val2014_000000341118_couch/f_sec_rtv}&
        & \includegraphics[valign = c, width=0.32\linewidth]{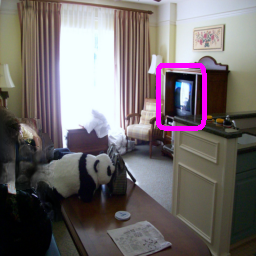}\\

         \bf Regular       & $S(\text{\it couch}) = \textcolor{red}{0.63}$\textcolor{red}{\Lightning}      &\multirow{2}{*}{\large$\boldmath\geq$ }& $S(\text{\it couch}) = \textcolor{red}{2.09}$\textcolor{red}{\Lightning}      \\ 
         \bf Ours & $S(\text{\it couch}) = \textcolor{PineGreen}{0.73}$&& $S(\text{\it couch}) = \textcolor{PineGreen}{-0.02}$\\[0.5em]

          \includegraphics[valign = c, width=0.9\linewidth]{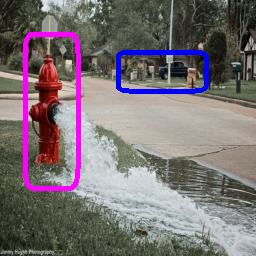} Original
        & \includegraphics[valign = c, width=0.32\linewidth]{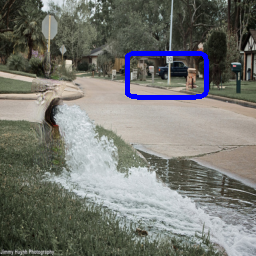}&
        & \includegraphics[valign = c, width=0.32\linewidth]{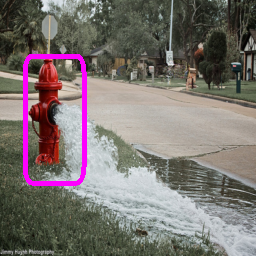}\\

         \bf Regular       & $S(\text{\it car}) = \textcolor{red}{-1.13}$\textcolor{red}{\Lightning}      &\multirow{2}{*}{\large$\boldmath\geq$ }& $S(\text{\it car}) = \textcolor{red}{0.08}$\textcolor{red}{\Lightning}      \\ 
         \bf Ours & $S(\text{\it car}) = \textcolor{PineGreen}{-0.96}$&& $S(\text{\it car}) = \textcolor{PineGreen}{-2.46}$\\[0.5em]
    \end{tabular}
    }
    \vspace*{-2mm}
    \caption{Context violations by image-level classifier. The primary object is marked with blue box and the context object is marked with magenta. The first column shows the original image, middle shows the image with only object and the third with only the context. We see that the baseline classifier depends heavily on the context and always scores the context only images~(last column) higher than the image with only the primary object (middle column). The data augmented model does better and gets the ordering right. }
    \vspace*{-1em}
    \label{fig:classInconsistapp}
\end{figure}

\myparagraph{Relation of performance to robustness}
As discussed in section $4.1.2$ in the main paper, we find that many well-performing object classes in terms of average precision~(AP) perform poorly in terms of robustness. To show this we plot the per-class average precision against the worst-case robustness metric $V^{\text{min}}(c_i)$ in \figref{fig:PerclassViol}. We can see for example that classes like mouse, tennis racket, sports ball, baseball bat and book which have high AP~($geq$ 0.6) have poor robustness ~( $V^{\text{min}}(c_i)$ > 0.5). In all these cases, the classifier seems to predominantly use contextual objects to make their predictions and achieve high average precision. But they fail when presented with object-without-context and context-without-object images, usually scoring the context-without-object images higher. This is also seen in further visual examples presented in \figref{fig:classInconsistapp}. Interestingly visually distinct classes like zebra, elephant, giraffe achieve high AP, while also being robust as seen in \figref{fig:PerclassViol} .

\subsection{Semantic Segmentation}
\label{sec:robusSegm}
\begin{figure}[t]
\centering
    \includegraphics[trim={1cm 0 1cm 0},width=1.0\linewidth]{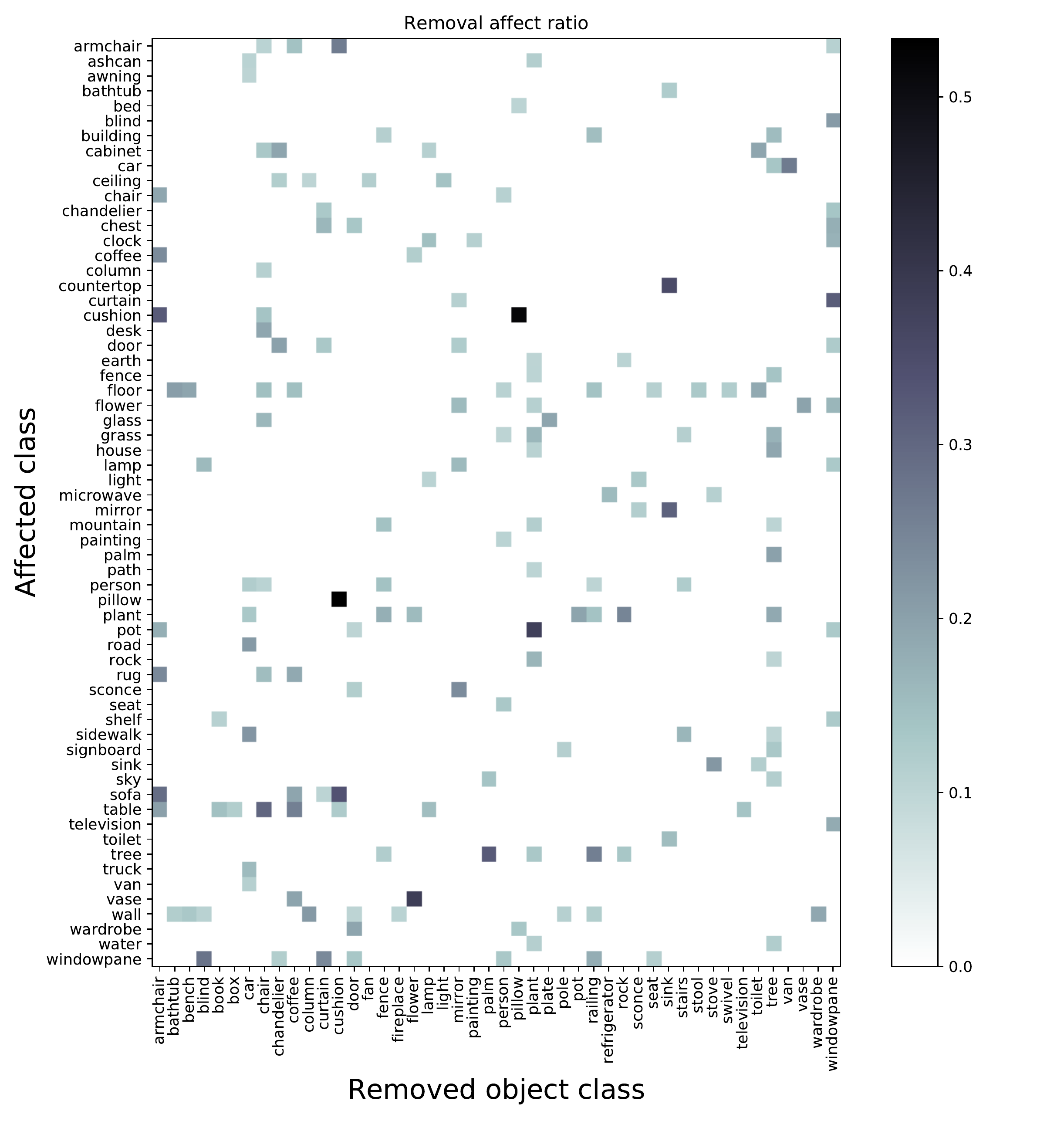}
    \caption{Visualizing frequency with which classes are affected by removal of other objects. Y-axis are the affected objects and the x-axis shows the removed objects}
\label{fig:segmAffect}
\end{figure}

\myparagraph{Visualizing robustness metrics} We compute the robustness metric $AR(c_i,c_j)$, which measures the ratio of instances when segmentation of class $c_i$ is affected by the removal of class $c_j$, in the ADE20k dataset for the Upernet~\cite{xiao2018unified} model.
This is visualized in \figref{fig:segmAffect}.  The y-axis the affected class and the x-axis is the removed object class. We show the rows and the columns which have atleast one entry $> 0.1$, for readability.
We can see that the $AR(c_i,c_j)$ matrix is very sparse, indicating that the segmentation is not affected by all removal, but of only specific classes. As discussed in section $4.2.2$ of the main paper, we can see that classes like road and sidewalk depend on the class car. The sidewalk is also to an extent affected by removal of trees. 

\begin{figure}[t]
\centering
    \includegraphics[trim={0.5cm 0 1cm 0},height=1.2\linewidth,width=1.0\linewidth]{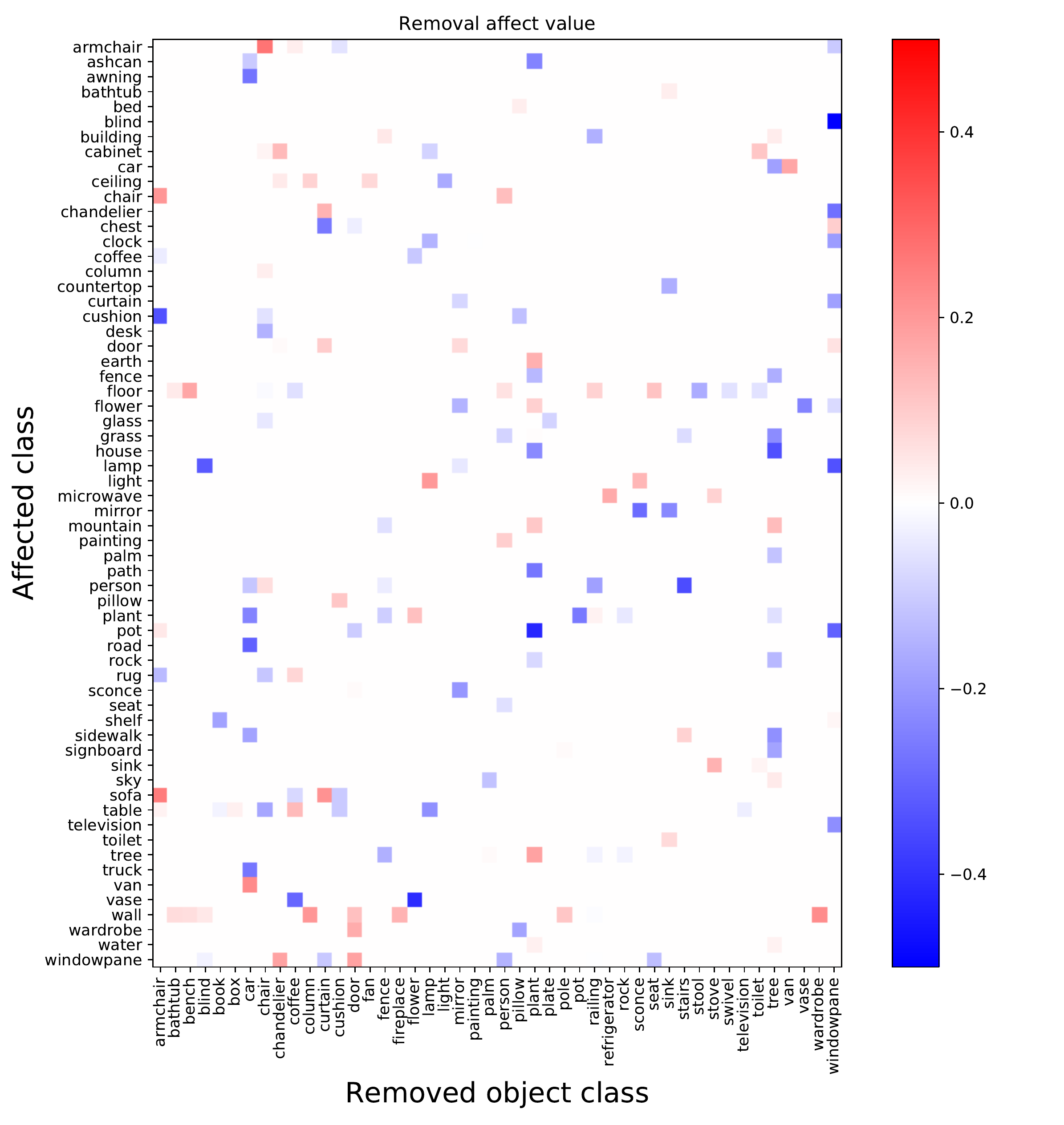}
    \caption{Visualizing the mean change in the IoU of object segmentation with context object removal. Y-axis are the affected objects and the x-axis shows the removed objects}
\label{fig:segmAffectValue}
\end{figure}

To measure the direction of the effect, that is if removal of context harms or improves the segmentation of a class, we visualize the average change in IoU in \figref{fig:segmAffectValue}.  Surprisingly, we find that not all context removal negatively affects the segmentation. Sometimes removing an object helps the model to resolve ambiguities and fix the segmentation of other objects. We can see in \figref{fig:segmAffectValue} that while majority of change is negative, for a few pairs of objects removal positively affects the IoU. For example removing \emph{lamp} class improves the segmentation of ceiling \emph{light}. Similarly, removing \emph{armchair} improves segmentation of \emph{chair} and \emph{sofa} classes, since the ambiguity is resolved.


\myparagraph{Ablation on data augmentation} To understand if removal and in-painting is really needed for data augmentation, we conduct two ablation studies which are presented in \Tableref{tab:da_segmentation_abl}. First we train a version of the baseline model where for each training sample we randomly select an object and set its label to 'ignore'. We use the same sampling strategy as \emph{sizebased} data augmentation model. Hence this mimics exactly the training procedure in DA (sizebased), except without actually removing the object. Comparing the results of this model (No removal (sizebased)) to the data augmented version, we see that removing the object is necessary and simply ignoring the label leads to a severe drop in performance (0.354 vs 0.379). Similarly, in the second experiment we train a model with the object removed but without in-painting. In this case we can see from \tableref{tab:da_segmentation_abl} that, having in-painter during augmentation is slightly better than the model without ~(0.379 vs 0.375).

Further visual examples of the sensitivity of the baseline Upernet~\cite{xiao2018unified} model to contextual changes and the robustness provided by data-augmentation is seen in \figref{fig:InconsistentSegApp}.

\myparagraph{Confirming the source of sensitivity} Finally we conduct an additional experiment to verify that the volatility we see in the output of the segmentation models on edited images are due to removal of context objects and not due to editing artifacts. To test this we observe the predictions of the model on three version of the input image. First is the original image. Second is the image with a context object removed. Finally the last image is the false edited image created by masking and in-painting the input image with the same mask as the context object, except with a horizontal flip. Thus the context object is not removed in the false edited image, but a similar shape and size region is removed and in-painted in a different part of the input image. All three images are fed to the segmentation models and the output is shown in \figref{fig:TestingSource}. We can see that the Upernet model output is virtually identical on the original image and the false edited image(third row). However the segmentation on the edited image with context object removed is significantly different (second row). This indicates that the segmentation models are not affected by the editing artifacts but by the removal of the context objects as claimed in the paper.

\begin{table}[t]
    \newcommand*{\mrowt}[1]{\multirow{2}{*}{#1}}
    \footnotesize
    \setlength{\tabcolsep}{1mm}
    \centering
    \begin{tabular}{lc cc}
    \toprule
    \multirow{2}{*}{Model} & Removed &  \multicolumn{2}{c}{ADE20k}\\\cmidrule(lr){3-4}
              & Pixels & mIoU & Acc\\\cmidrule(lr){1-2}\cmidrule(lr){3-4}
        Upernet\cite{xiao2018unified}      & - & 0.377 & 78.31 \\\cmidrule(lr){1-2}\cmidrule{3-4}
        No removal (sizebased) & Ignore & 0.354 & 77.45 \\
        No inpainter (sizebased) & Ignore & 0.375 & 78.25 \\
        DA (sizebased)     & Ignore   & 0.379 & 78.31 \\
     \bottomrule
    \end{tabular}
    \vspace*{-1mm}
    \caption{Data augmentation results on ADE20k dataset}
    \vspace*{-5mm}
    \label{tab:da_segmentation_abl}
\end{table}
\begin{figure*}
    \setlength{\tabcolsep}{1pt}
    \noindent\makebox[\textwidth]{
    \scriptsize
    \begin{tabular}{cccccc}
          \includegraphics[valign = c, height=0.06\linewidth, width=0.16\linewidth]{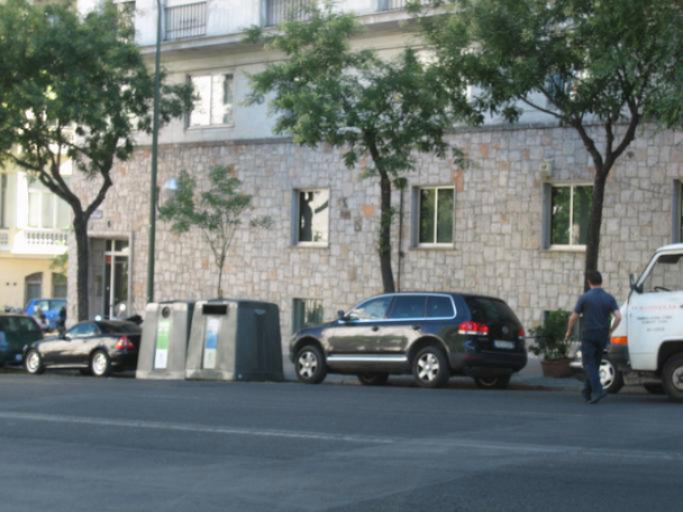}
        & \includegraphics[valign = c, height=0.06\linewidth, width=0.16\linewidth]{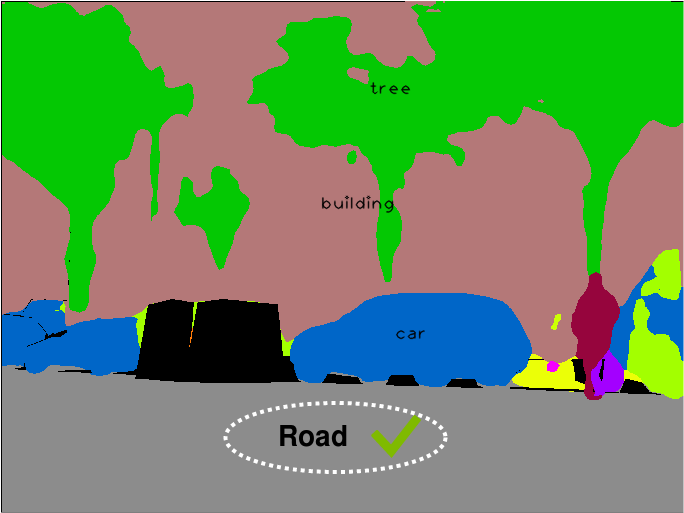} 
        & \includegraphics[valign = c, height=0.06\linewidth, width=0.16\linewidth]{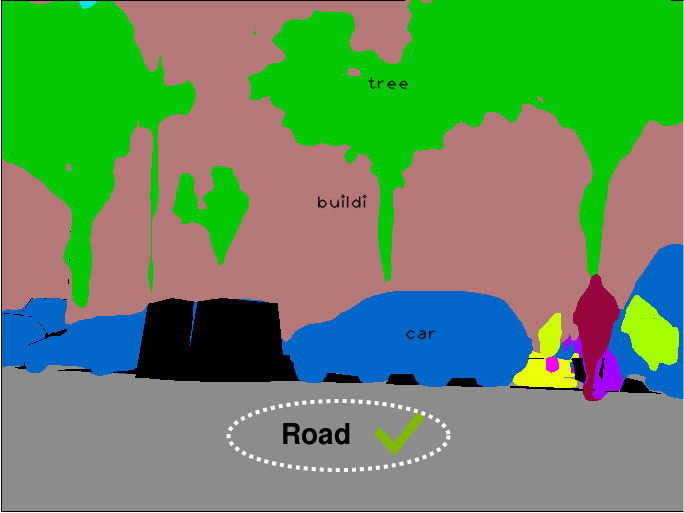} 
        & \includegraphics[valign = c, height=0.06\linewidth, width=0.16\linewidth]{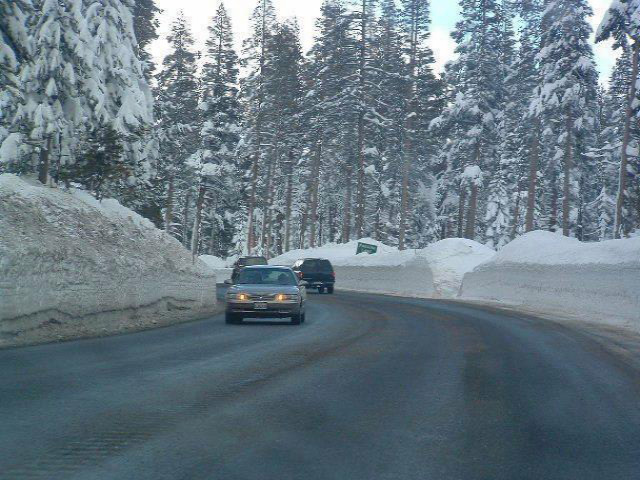}
        & \includegraphics[valign = c, height=0.06\linewidth, width=0.16\linewidth]{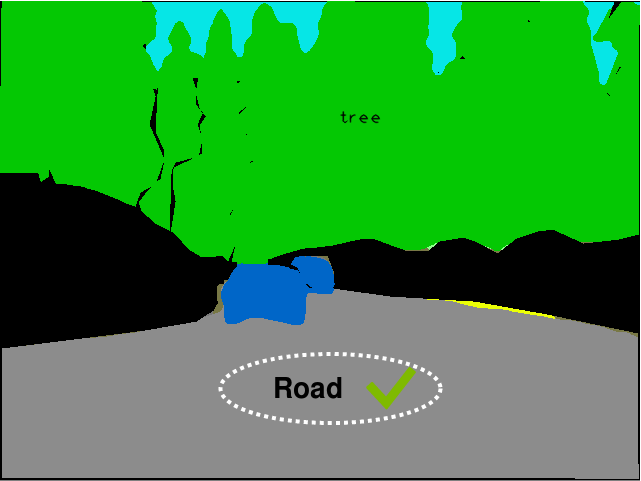}
        & \includegraphics[valign = c, height=0.06\linewidth, width=0.16\linewidth]{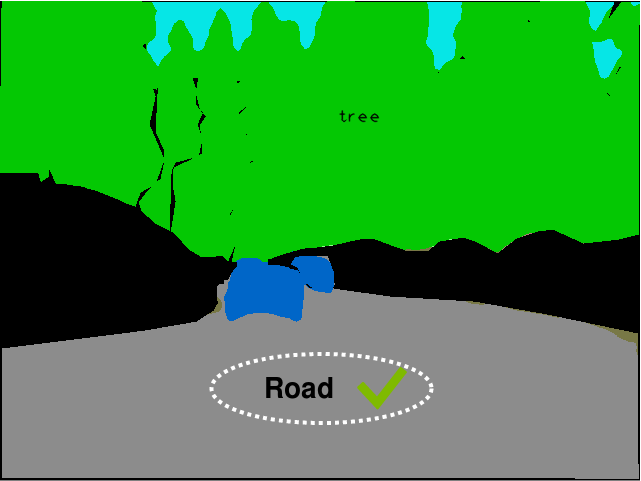}\\
          original $(\mathcal{I})$ & Upernet   & Ours & original $(\mathcal{I})$ & Upernet & Ours\\
          \includegraphics[valign = c, height=0.06\linewidth, width=0.16\linewidth]{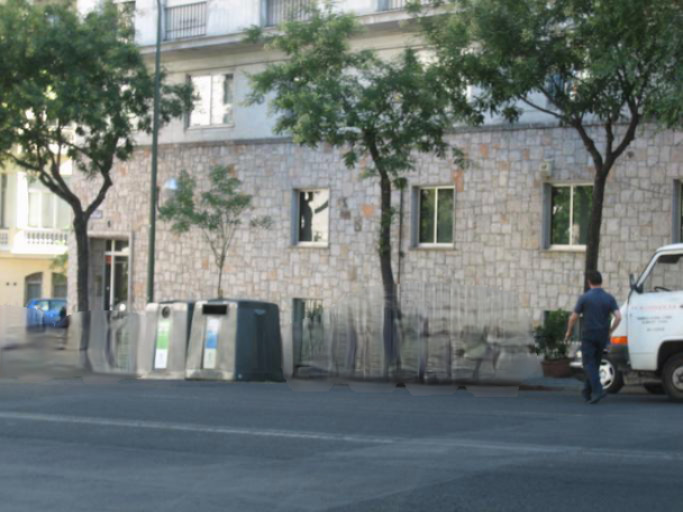}
        & \includegraphics[valign = c, height=0.06\linewidth, width=0.16\linewidth]{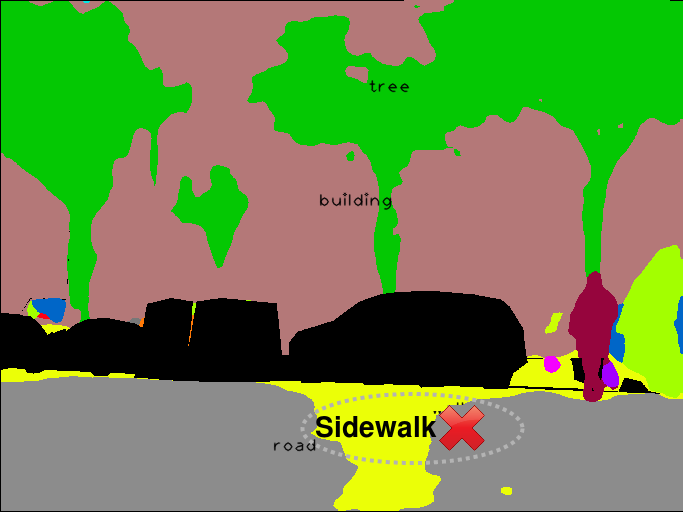}
        & \includegraphics[valign = c, height=0.06\linewidth, width=0.16\linewidth]{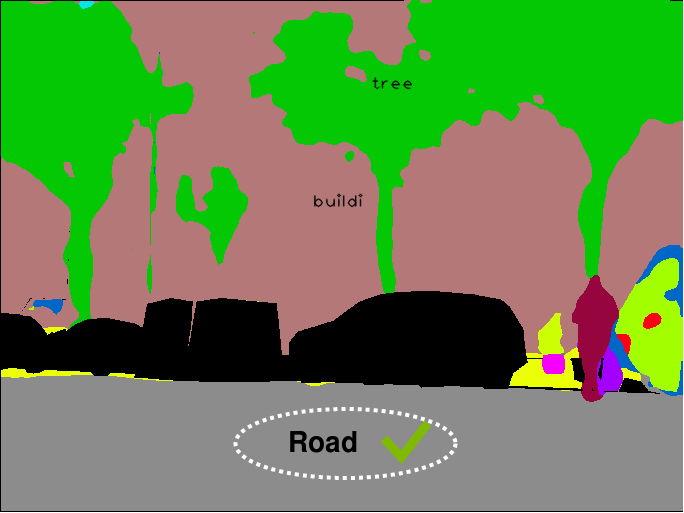}
        & \includegraphics[valign = c, height=0.06\linewidth, width=0.16\linewidth]{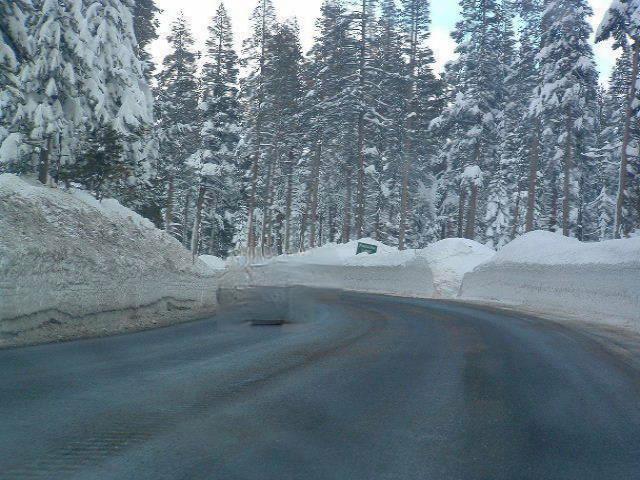}
        & \includegraphics[valign = c, height=0.06\linewidth, width=0.16\linewidth]{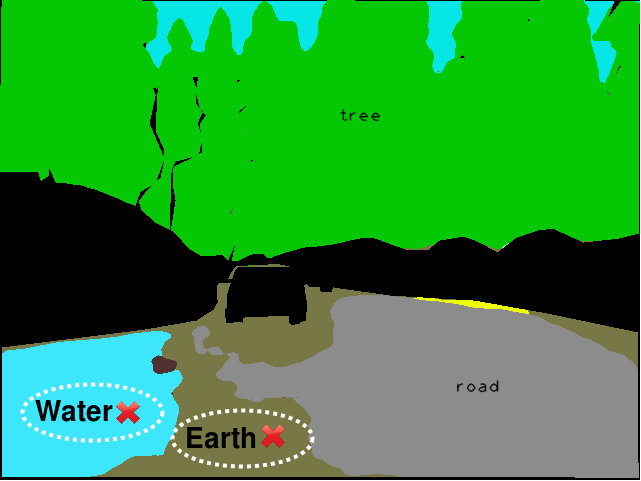}
        & \includegraphics[valign = c, height=0.06\linewidth, width=0.16\linewidth]{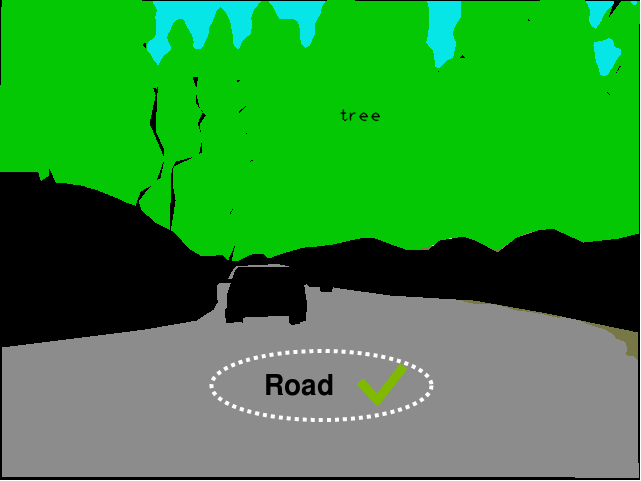}\\
          $\mathcal{I} - car$ & Upernet & Ours & $\mathcal{I} - car$ & Upernet & Ours\\[0.3em]
          \includegraphics[valign = c, height=0.06\linewidth, width=0.16\linewidth]{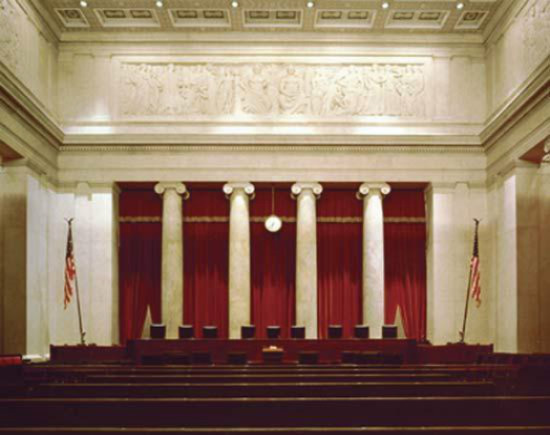}
        & \includegraphics[valign = c, height=0.06\linewidth, width=0.16\linewidth]{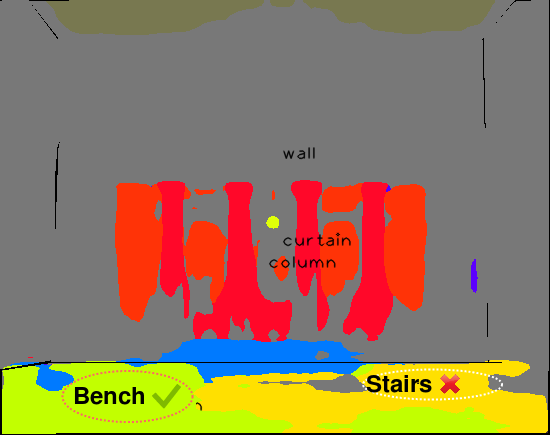} 
        & \includegraphics[valign = c, height=0.06\linewidth, width=0.16\linewidth]{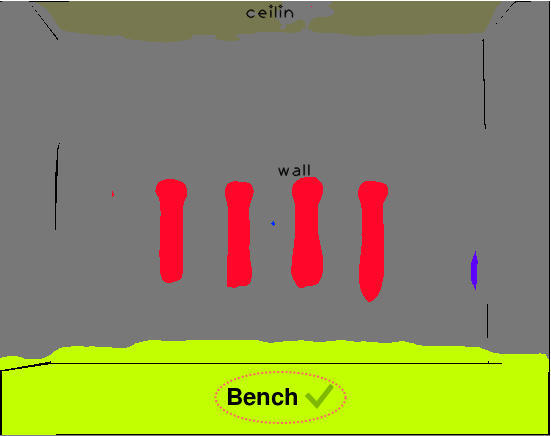} 
        & \includegraphics[valign = c, height=0.06\linewidth, width=0.16\linewidth]{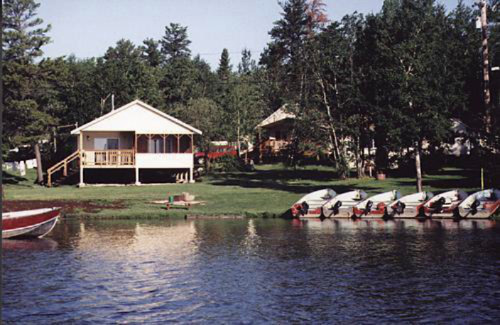}
        & \includegraphics[valign = c, height=0.06\linewidth, width=0.16\linewidth]{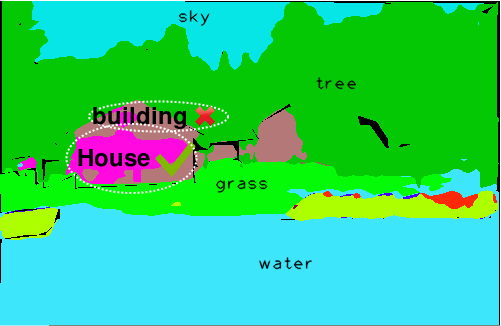}
        & \includegraphics[valign = c, height=0.06\linewidth, width=0.16\linewidth]{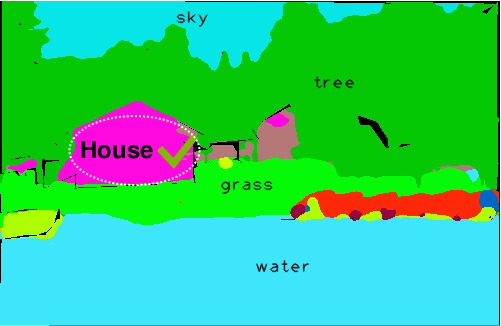}\\
          original $(\mathcal{I})$ & Upernet  & Ours & original $(\mathcal{I})$ & Upernet & Ours\\
          \includegraphics[valign = c, height=0.06\linewidth, width=0.16\linewidth]{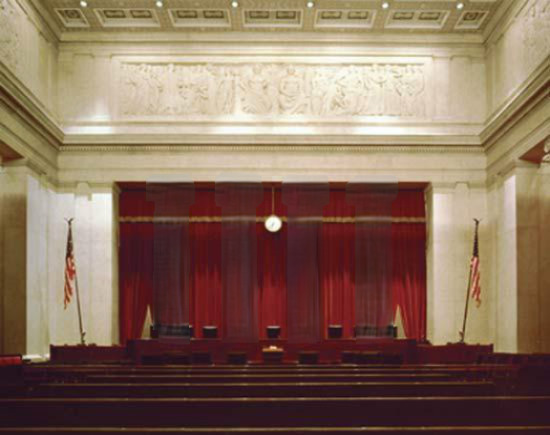}
        & \includegraphics[valign = c, height=0.06\linewidth, width=0.16\linewidth]{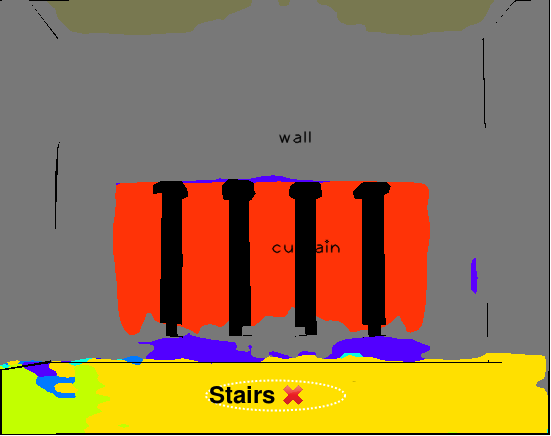} 
        & \includegraphics[valign = c, height=0.06\linewidth, width=0.16\linewidth]{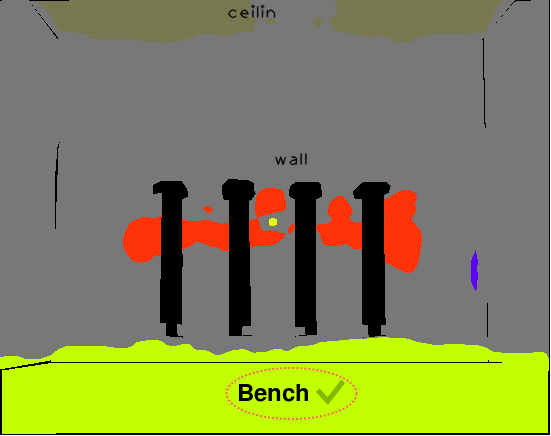} 
        & \includegraphics[valign = c, height=0.06\linewidth, width=0.16\linewidth]{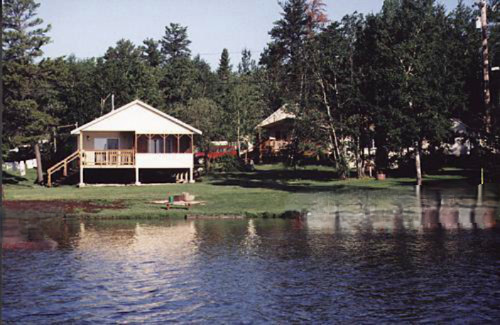}
        & \includegraphics[valign = c, height=0.06\linewidth, width=0.16\linewidth]{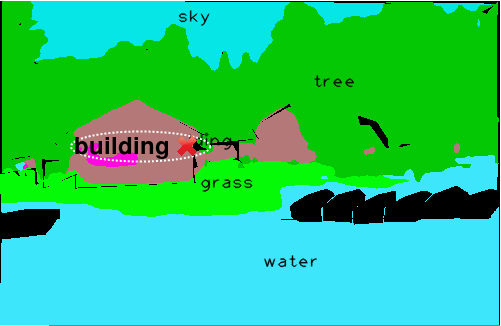}
        & \includegraphics[valign = c, height=0.06\linewidth, width=0.16\linewidth]{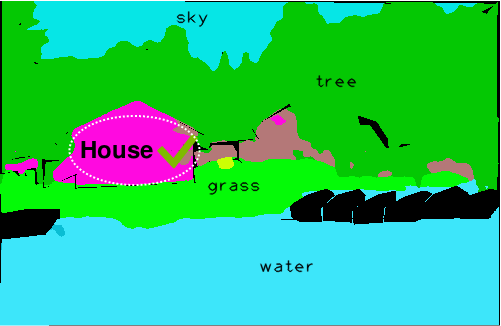}\\
          $\mathcal{I} - column$ & Upernet  & Ours & $\mathcal{I} - boat$ & Upernet & Ours\\
    \end{tabular}
        }
    \vspace*{-0.5em}
    \caption{Examples of segmentation failures due to removal of a single context object. We see the segmentation of road, bench and house affected significantly when context objects like car, columns and boat is removed~(comparing odd and even rows). Model trained with proposed data-augmentation is more robust to these changes.}
    \vspace*{-1em}
    \label{fig:InconsistentSegApp}
\end{figure*}

\begin{figure*}
    \setlength{\tabcolsep}{1pt}
    \noindent\makebox[\textwidth]{
    \scriptsize
    \begin{tabular}{cccccc}
          \includegraphics[valign = c, height=0.06\linewidth, width=0.16\linewidth]{segmentation/correlErrors/ADE_val_00001804_road/original}
        & \includegraphics[valign = c, height=0.06\linewidth, width=0.16\linewidth]{segmentation/correlErrors/ADE_val_00001804_road/pred_prim}
        & \includegraphics[valign = c, height=0.06\linewidth, width=0.16\linewidth]{segmentation/correlErrors/ADE_val_00001804_road/pred_sec}
        & \includegraphics[valign = c, height=0.06\linewidth, width=0.16\linewidth]{segmentation/correlErrors/ADE_val_00000819_sidewalk/original}
        & \includegraphics[valign = c, height=0.06\linewidth, width=0.16\linewidth]{segmentation/correlErrors/ADE_val_00000819_sidewalk/pred_prim} 
        & \includegraphics[valign = c, height=0.06\linewidth, width=0.16\linewidth]{segmentation/correlErrors/ADE_val_00000819_sidewalk/pred_sec}\\ 
          original $(\mathcal{I})$ & Upernet   & Ours & original $(\mathcal{I})$ & Upernet & Ours\\[0.3em]
          \includegraphics[valign = c, height=0.06\linewidth, width=0.16\linewidth]{segmentation/correlErrors/ADE_val_00001804_road/original_edited}
        & \includegraphics[valign = c, height=0.06\linewidth, width=0.16\linewidth]{segmentation/correlErrors/ADE_val_00001804_road/pred_prim_edited}
        & \includegraphics[valign = c, height=0.06\linewidth, width=0.16\linewidth]{segmentation/correlErrors/ADE_val_00001804_road/pred_sec_edited}
        & \includegraphics[valign = c, height=0.06\linewidth, width=0.16\linewidth]{segmentation/correlErrors/ADE_val_00000819_sidewalk/original_edited}
        & \includegraphics[valign = c, height=0.06\linewidth, width=0.16\linewidth]{segmentation/correlErrors/ADE_val_00000819_sidewalk/pred_prim_edited}
        & \includegraphics[valign = c, height=0.06\linewidth, width=0.16\linewidth]{segmentation/correlErrors/ADE_val_00000819_sidewalk/pred_sec_edited}\\
          $\mathcal{I} - car$ & Upernet & Ours & $\mathcal{I} - sign$ & Upernet & Ours\\[0.3em]
          \includegraphics[valign = c, height=0.06\linewidth, width=0.16\linewidth]{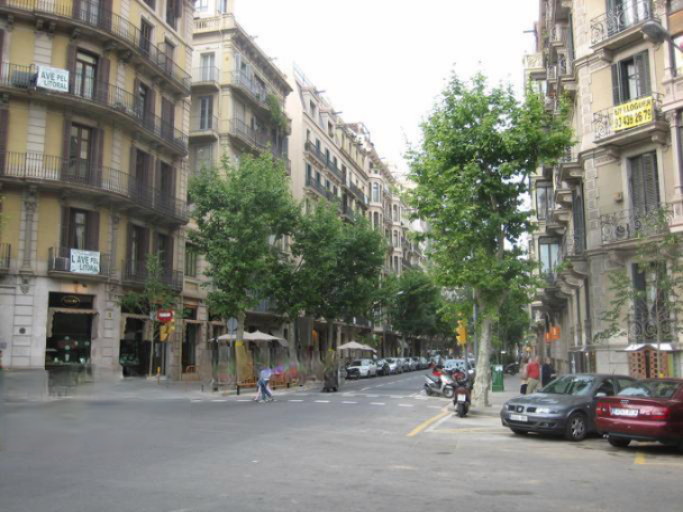}
        & \includegraphics[valign = c, height=0.06\linewidth, width=0.16\linewidth]{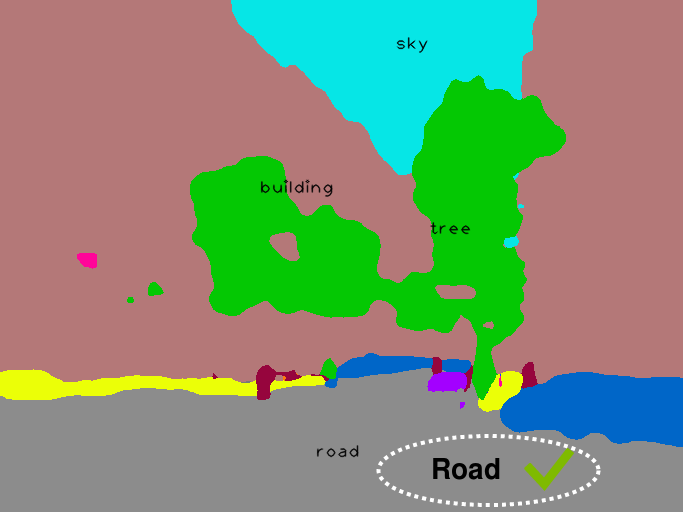}
        & \includegraphics[valign = c, height=0.06\linewidth, width=0.16\linewidth]{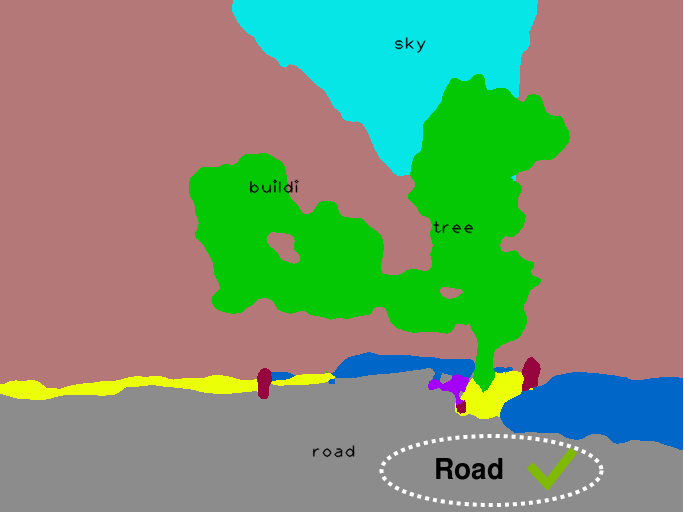}
        & \includegraphics[valign = c, height=0.06\linewidth, width=0.16\linewidth]{segmentation/dummyErrors/ADE_val_00000819_sidewalk_signboard/original_edited}
        & \includegraphics[valign = c, height=0.06\linewidth, width=0.16\linewidth]{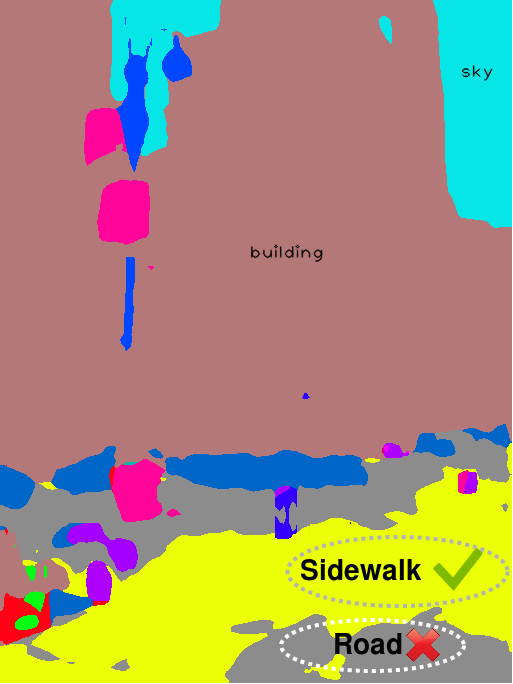}
        & \includegraphics[valign = c, height=0.06\linewidth, width=0.16\linewidth]{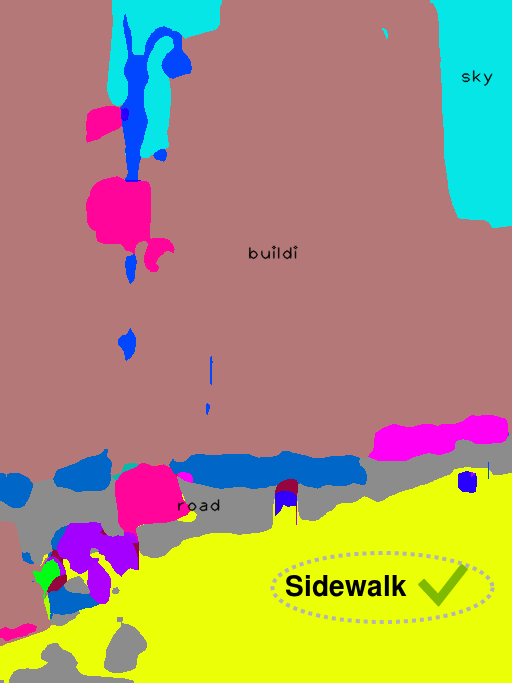}\\
          $\mathcal{I} - car$ (flipped mask) & Upernet & Ours & $\mathcal{I} - sign$ (flipped mask) & Upernet & Ours\\[0.3em]
    \end{tabular}
        }
    \vspace*{-0.5em}
    \caption{Experiment to verify that the volatality of the segmentation output is due to object removal and not due to editing artifacts. First row shows original images and the segmenations produced for them. Second row shows the edited images with an object removed and the segmentation output for them. Here we can see the segmentation output of Upernet significantly affected by the removal of car and sign. Final, row shows the original image edited with the same object mask as the second row, but horizontally flipped. Thus the object is not removed, but a different part of the image is edited with the same mask. We can see here that the segmentation is not affected at all by this edit and is very similar to the segmentation produced by the original image.}
    \vspace*{-1em}
    \label{fig:TestingSource}
\end{figure*}